\documentclass[journal]{IEEEtran}
\usepackage{cite}
\usepackage{times}
\usepackage{epsfig}
\usepackage{graphicx}
\usepackage{amsmath}
\usepackage{amssymb}
\usepackage{algorithm, algorithmic}
\usepackage{enumerate}
\usepackage{multirow}
\usepackage{xcolor}
\usepackage{graphics}
\usepackage{epsfig}
\usepackage{epstopdf}
\usepackage[switch]{lineno}
\usepackage{url}

\makeatletter
\def\hlinew#1{%
  \noalign{\ifnum0=`}\fi\hrule \@height #1 \futurelet
   \reserved@a\@xhline}
\makeatother

\begin{document}

\title{Improved Deep Hashing with Soft Pairwise Similarity for Multi-label Image Retrieval}

\author{Zheng~Zhang,
        Qin~Zou,~\IEEEmembership{Senior Member,~IEEE},
        Yuewei~Lin,
        Long Chen, \\
        and Song Wang,~\IEEEmembership{Senior Member,~IEEE}
\thanks{This research was supported by the National Natural Science
Foundation of China under grant 61872277, 41571437, 61672376, and U1803264, the Hubei Provincial Natural
Science Foundation under grant 2018CFB482. Yuewei Lin gratefully acknowledges the support by BNL LDRD 18-009. (Corresponding author: Qin Zou)
}
\thanks{Z.~Zhang and Q.~Zou are with the
School of Computer Science, Wuhan University, Wuhan 430072,
P.R.~China (E-mails: \{zhengzhang, qzou\}@whu.edu.cn).}
\thanks {Y.~Lin is with the Computational Science Initiative, Brookhaven National Laboratory, NY 11973, USA (E-mail: ywlin@bnl.gov).}
\thanks{L.~Chen is with the School of Data and Computer Science, Sun Yat-Sen University, Guangzhou 518001,
P.R.~China (E-mail: chenl46@mail.sysu.edu.cn).}
\thanks {S.~Wang is with the Department of Computer Science and Engineering,
University of South Carolina, Columbia, SC 29201 USA, and also with the
School of Computer Science and Technology, Tianjin University, Tianjin
300072, China (E-mail: songwang@cec.sc.edu).}
%\thanks{Manuscript received Dec ??, 2015.}
}

%\markboth{IEEE TIFS} {Shell \MakeLowercase{\textit{et al.}}: }
\markboth{IEEE Transactions on Multimedia, 2019}
{Shell \MakeLowercase{\textit{et al.}}: }


\maketitle
%\linenumbers

%with Quantitative Pairwise Similarity and Joint Losses

\begin{abstract}
Hash coding has been widely used in the approximate nearest neighbor search for large-scale image retrieval. Recently, many deep hashing methods have been proposed and shown largely improved performance over traditional feature-learning methods. Most of these methods examine the pairwise similarity on the semantic-level labels, where the pairwise similarity is generally defined in a hard-assignment way. That is, the pairwise similarity is `1' if they share no less than one class label and `0' if they do not share any. However, such similarity definition cannot reflect the similarity ranking for pairwise images that hold multiple labels. In this paper, an improved deep hashing method is proposed to enhance the ability of multi-label image retrieval. We introduce a pairwise quantified similarity calculated on the normalized semantic labels. Based on this, we divide the pairwise similarity into two situations -- `hard similarity' and `soft similarity', where cross-entropy loss and mean square error loss are adapted respectively for more robust feature learning and hash coding. Experiments on four popular datasets demonstrate that, the proposed method outperforms the competing methods and achieves the state-of-the-art performance in multi-label image retrieval.
\end{abstract}


\IEEEpeerreviewmaketitle

\section{Introduction} \label{sec:intro}
With the popular use of smartphone cameras, the amount of image data has been rapidly increasing, which calls for more efficient and accurate image retrieval. Generally, image retrieval is based on the approximate nearest neighbor search~\cite{tao2006direct} and an image-retrieval system is often built on hashing~\cite{wang2014hashing}. In hashing methods, high dimensional data are transformed into compact binary codes and similar binary codes are expected to generate for similar data items. Due to the encouraging efficiency in both speed and storage, a number of hashing methods have been proposed in the past decade~\cite{weiss2009spectral,kulis2009learning,wang2010semi,norouzi2011minimal,liu2012supervised,gong2013iterative,Li2015Ordinal,Liu2016Sequential,Jie2016Supervised,Liu2017Reversed,shen2018unsupervised,zhou2018graph}.

Generally, the existing hashing methods can be divided into two categories: unsupervised methods and supervised methods. The unsupervised methods use unlabeled data to generate hash functions. They focus on  preserving the distance similarity in the Hamming space as in the feature space. The supervised methods incorporate human-interactive annotations, e.g., pairwise similarities of semantic labels, into the learning process to improve the quality of hashing, and often outperform the unsupervised methods. In the past five years, inspired by the success of deep neural networks that show superior feature-representation power in image classification \cite{krizhevsky2012imagenet,simonyan2014very,szegedy2015going,grsl2015}, object detection \cite{szegedy2013deep}, face recognition \cite{sun2014deep}, and many other vision tasks~\cite{long2015fully,deng2014large,Zou2018deepcrack}, many supervised hashing methods based on deep neural networks were developed for image abstraction and hash-code learning~\cite{xia2014supervised,zhao2015deep,lai2015simultaneous,zhang2015bit,zhu2016deep,cao2016deep,liu2016deep,cao2017hashnet,li2017deep,wang2016deep,li2016feature,jiang2017deep,he2018hashing}. These so called deep hashing methods have achieved the state-of-the-art performance on several popular benchmark datasets.

\begin{figure}[!t]
  \centering
  \includegraphics[width=1.0\linewidth]{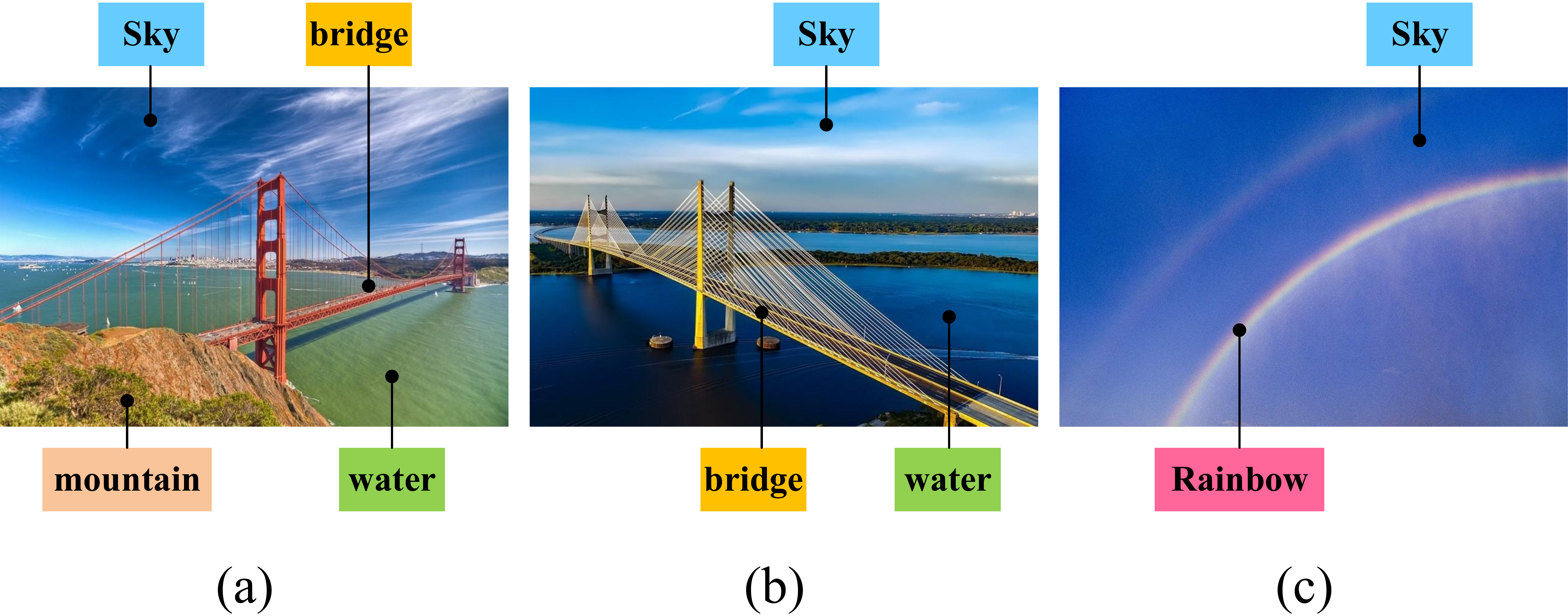}\vspace{-2mm}
  \caption{Some examples of the multi-label images. Since the images in (a) and (b) share more class labels, the similarity between them is supposed to be higher than that between (a) and (c). However, in the traditional pairwise-similarity definition, the similarities between them are the same.}
  \label{fig:one}
\end{figure}

While these supervised deep hashing methods have produced impressive improvement in image retrieval, to the best of our knowledge, they only examine the similarity of pairwise images using the semantic-level labels, and define the similarity in a coarse way. That is, {\it the similarity of pairwise images is `1' if they share at least one object class and `0' (or `-1') if they do not share any object class}. However, such similarity definition cannot reflect the fine-grained similarity when the pairwise images both have multiple labels. An illustrative example is shown in Fig.~\ref{fig:one} where the images in (a), (b) and (c) share the same class label `sky' and each pair of them are taken as similar in the context of image retrieval. However, as the images in (a) and (b) share three class labels, i.e., `sky', `bridge', and `water', the similarity between them should be ranked higher than that between (a) and (c) which have only one class label in common.
It can be easily observed that, the traditional coarse similarity definition does not take the multi-label information into account and cannot rank the similarity for images with multiple class labels.

To solve this problem, we present a soft definition for the pairwise similarity with regarding to the semantic labels each image holds. Specifically, the pairwise similarity is quantified into a percentage using the normalized semantic labels. Based on the quantified similarity, we propose a deep method to improve the retrieval quality for multi-label image retrieval. For convenience, we abbreviate this improved deep hashing network as IDHN in the following description. Specifically, we divide the quantified similarity into two situations: one is `hard similarity', which means a pair of images share either all object types or none; another is `soft similarity', which means a pair of images share some object classes, but not all. For robustness and practicability, we construct cross entropy loss for `hard similarity' situation and mean square error loss for `soft similarity', for preserving the similarity of an image pair in hash space converging to their fine-grained semantic similarity in form of normalized labels. We evaluate the proposed deep hashing method on four popular multi-label image datasets and obtain significantly improved performance over the state-of-the-art hashing methods in image retrieval. The contributions of this work lie in three-fold:

\begin{enumerate}[$\vcenter{\hbox{\small$\bullet$}}$]
\item We propose a soft definition for the pairwise similarity by quantifying it into a percentage using the normalized semantic labels. To the best of our knowledge, IDHN is the first deep hashing method that directly uses pairwise quantified similarity which can reflect the fine-grained similarity between a pair of multi-label images for supervised learning.
%\vspace{-1mm}

\item We divide the pairwise similarity into two situations -- `hard similarity' and `soft similarity', and a joint loss-function of cross-entropy loss and mean square error loss are adapted for learning efficient, robust hash codes, and preserving the fine-grained semantic similarity based on the quantified similarity.
%\vspace{-1mm}

\item Experiments have shown that the proposed method outperforms current state-of-the-art methods on four datasets in image retrieval and has good extensibility to deeper network architecture, which demonstrates the effectiveness of the proposed method.

\end{enumerate}

The rest of this paper is organized as follows: Section~\ref{sec:relate} briefly reviews the related work. Section~\ref{sec:model} describes the proposed quantified similarity deep hashing method which generates high-quality hash codes in a supervised learning manner. Section~\ref{sec:experiment} demonstrates the effectiveness of the proposed model by extensive experiments on four popular benchmark datasets, and Section~\ref{sec:conc} concludes our work.

\section{Related Work}\label{sec:relate}
In the past two decades, many hashing methods have been proposed for approximate nearest neighbor search in the large-scale image retrieval. Hashing-based methods transform high dimensional data into compact binary codes with a fixed number of bits and generate similar binary codes for similar data items, which can greatly reduces the storage and calculation consumption. Generally, the existing hashing methods can be divided into two categories: unsupervised methods and supervised methods.

\textbf{Unsupervised Methods.} The unsupervised hashing methods learn hash functions to preserve the similarity distance in the Hamming space as in the feature space. Locality-Sensitive Hashing (LSH)~\cite{datar2004locality} is one of the most well-known representative. LSH aims to maximize the probability that the similar items will be mapped to the same buckets. Spectral Hashing (SH)~\cite{weiss2009spectral} and~\cite{li2013spectral} consider hash encoding as a spectral graph partitioning problem, and learn a nonlinear mapping to preserve semantic similarity of the original data in the Hamming space. Iterative Quantization (ITQ)~\cite{gong2013iterative} searches for an orthogonal matrix by alternating optimization to learn the hash functions. Sparse Product Quantization (SPQ)~\cite{ning2017scalable} encodes the high-dimensional feature vectors into sparse representation by decomposing the feature space into a Cartesian product of low-dimensional subspaces and quantizing each subspace via K-means clustering, and the sparse representations are optimized by minimizing their quantization errors. \cite{ercoli2017compact} proposes to learn compact hash code by computing a sort of soft assignment within the k-means framework, which is called "multi-k-means", to void the expensive memory and computing requirements. Latent Semantic Minimal Hashing (LSMH)~\cite{lu2016latent} refines latent semantic feature embedding in the image feature to refine original feature based on matrix decomposition, and a minimum encoding loss is combined with latent semantic feature learning process simultaneously to get discriminative obtained binary codes.

\textbf{Supervised Methods.} The supervised hashing methods use supervised information to learn compact hash codes, which usually achieve superior performance compared with the unsupervised methods. Binary Reconstruction Embedding (BRE)~\cite{kulis2009learning} constructs hash functions by minimizing the squared error loss between the original feature distances and the reconstructed Hamming distances. Semi-supervised hashing~(SSH) \cite{wang2010semi} combines the characteristics of the labeled and unlabeled data to learning hash functions, where the supervised term tries to minimize the empirical error on the labeled data and the unsupervised term pursuits effective regularization by maximizing the variance and independence of hash bits over the whole data. Minimal Loss Hashing (MLH)~\cite{norouzi2011minimal} learns hash functions based on structural prediction with latent variables using a hinge-like loss function. Supervised Hashing with Kernels (KSH)~\cite{liu2012supervised} is a kernel based method which learns compact binary codes by maximizing the separability between similar and dissimilar pairs in the Hamming space.
Online Hashing~\cite{Huang2017Online} is also a hot research area in image retrieval. ~\cite{Xia2013Online} proposes an online multiple kernel learning method, which aims to find the optimal combination of multiple kernels for similarity learning, and ~\cite{Liang2017Semisupervised} improves the online multi-kernel learning with semi-supervised way, which utilizes supervision information to estimate the labels of the unlabeled images by introducing classification confidence that is also instructive to select the reliably labeled images for training.

In the last few years, approaches built on deep neural networks have achieved state-of-the-art performance on many vision tasks~\cite{krizhevsky2012imagenet,simonyan2014very,szegedy2015going} as comparing to traditional methods~\cite{zou2017robust}. Inspired by the powerful representation ability of deep neural networks, some deep hashing methods have been proposed, which show great progress compared with traditional hand-crafted feature based methods. A simple way to deep hashing learning is thresholding high level feature directly, the typical methods is DLBHC~\cite{Lin2015Deep}, which learns hash-like representations by inserting a latent hash layer before the last classification layer in AlexNet~\cite{krizhevsky2012imagenet}. While the network is fine-tuned well on classification task, the feature of latent hash layer is considered to be discriminative, which indeed presents better performance than hand-crafted feature.
CNNH~\cite{xia2014supervised} was proposed as a two-stage hashing method, which decomposes the hash learning process into a stage of learning approximate hash codes, and followed by a stage of deep network fine-tune to learn the image features and hash functions. DNNH~\cite{lai2015simultaneous} improves the two-stage CNNH in both the image representations and hash coding by using a joint learning process. DNNH and DSRCH \cite{zhang2015bit} use image triplets as the input of deep network, which generate hash codes by minimizing the triplet ranking loss. Since the pairwise similarity is more straightforward than the triplet similarity, most of the latest deep hashing networks used pairwise labels for supervised hashing and further improved the performance of image retrieval, e.g., DHN~\cite{zhu2016deep}, DQN~\cite{cao2016deep} and DSH~\cite{liu2016deep} etc. HashNet~\cite{cao2017hashnet} proposes a deep hashing method to learn binary hash codes from imbalanced similarity data by continuation method with convergence guarantees.

Since the above-mentioned deep hashing methods are not designed for multi-label image retrieval, the fine-grained similarity of multi-label images are always neglected with coarse-grained definition between pair images.
For multi-label retrieval, DSRH~\cite{zhao2015deep} tries to learn hash function by utilizing the ranking information of multi-level similarity, and proposes a surrogate losses to solve the optimization problem of ranking measures.
IAH~\cite{lai2016instance} focuses on learning instance-aware image representations and using the weighted triplet loss to preserve similarity ranking for multi-label images.
However, the weighted triplet loss functions adapted by DSRH~\cite{zhao2015deep} and IAH~\cite{lai2016instance} do not enforce direct restriction to learn fine-grained multilevel semantic similarity, since they are focusing on preserving the correct ranking of images according to their similarity degrees to the queries, which make it a room for improvement on the accuracies of top returned images.
Based on this, DMSSPH~\cite{wu2017deep} tries to construct hash functions to maximize the discriminability of the output space to preserve multilevel similarity between multi-label images. Althought DMSSPH~\cite{wu2017deep} has utilized the fine-grained multilevel semantic similarity for pairwise similarity learning, there still are spaces for further exploration.
A novel and effective method TALR was proposed in~\cite{he2018hashing}, which considered tied rankings on integer-valued Hamming distance and directly optimized the ranking-based evaluation metrics Mean Average Precision (MAP)~\cite{baeza1999modern} and Normalized Discounted Cumulative Gains (NDCG)~\cite{jarvelin2002cumulated}. It achieved high performance in several benchmark datasets. In~\cite{sablayrolles2017icassp}, two new protocols were presented for the evaluation of supervised hashing methods, under the context of transfer learning.

In this work, we study to improve the hashing quality by exploring the diversities of pairwise semantic similarity on the multi-label dataset.
Specifically, we propose to define the fine-grained pairwise similarity in the form of continuous value, and according to this definition, we divide the pairwise similarity into two situations and construct a joint pairwise loss function to perform simultaneous feature learning and hash-code generating.

\section{Improved Deep Hashing Networks}
\label{sec:model}
\subsection{Problem Definition}
Given a training set of $N$ images $I=\{ I_{1},I_{2},\cdots ,I_{N} \}$ and a pairwise similarity matrix $S=\{ s_{ij}| i,j=1,2,...,N \}$, the goal of hash learning for images is to learn a mapping
$F : I \mapsto \{-1, 1 \}^{q}$, so that an input image $I_i$ can be encoded into a $q$-bit binary code $F(I_i)$, with the similarities of images being preserved.
The similarity label $s_{ij}$ is usually defined as $s_{ij}$ = 1 if $I_{i}$ and $I_{j}$ have semantic label, i.e., object class label, in common and $s_{ij}$ = 0 if $I_{i}$ and $I_{j}$ do not share any semantic label. As discussed in the introduction, this definitions does not take the multi-label information into account and cannot rank the similarity for images with multiple class labels. In our design, the pairwise similarity is quantified into percentages and the similarity value $s_{ij}$ is defined as the cosine distance of pairwise label vectors:
\begin{equation}
\label{eq:s}
{s}_{ij}=\frac{\langle l_{i},l_{j} \rangle}{ \|  l_{i}  \| \|  l_{j}  \|},
\end{equation}
where $l_{i}$ and $l_{j}$ denote the semantic label vector of image $I_{i}$ and $I_{j}$, respectively, and $\langle l_{i},l_{j} \rangle$ calculates the inner product.
This cosine distance has been widely adopted in retrieval system, but it is always used to measure similarity of the feature vectors~\cite{cao2016deep}. To the best of our knowledge, we are the first to use the cosine distance to quantify fine-grained semantic similarity of pair images.

According to Eq.~(\ref{eq:s}), the similarity of pairwise images can be passed into three states: completely similar, partially similar, and dissimilar. For approximate nearest neighbor search, we demand that the binary codes $B = \{b_i\}_{i=1}^{N}$ should preserve the similarity in $S$. To be specific, given a pair of binary codes $b_{i}$ and $b_{j}$, if $s_{ij}$ = 0 which means pairwise images $I_{i}$ and $I_{j}$ do not share any object class, the Hamming distance between $b_{i}$ and $b_{j}$ should be large, i.e., be close to $q$ in the $q$-bit hash coding case; if $s_{ij}$ = 1, which means the pairwise images $I_{i}$ and $I_{j}$ have the same class labels, we expect the Hamming distance to be zero; otherwise, the binary codes $b_{i}$ and $b_{j}$ should have a suitable Hamming distance complying with the soft definition of similarity $s_{ij}$.

As a conclusion, we define the completely similar and dissimilar situation as `hard similarity', which can be seen as equivalent to the similarity definition for single label images. Besides of these `hard similarity' situations, the similarity between a pair of images is more fine and complicated, which we define as `soft similarity'.

\begin{figure*}[!t]
  \centering
  \centerline{\includegraphics[width=0.96\linewidth]{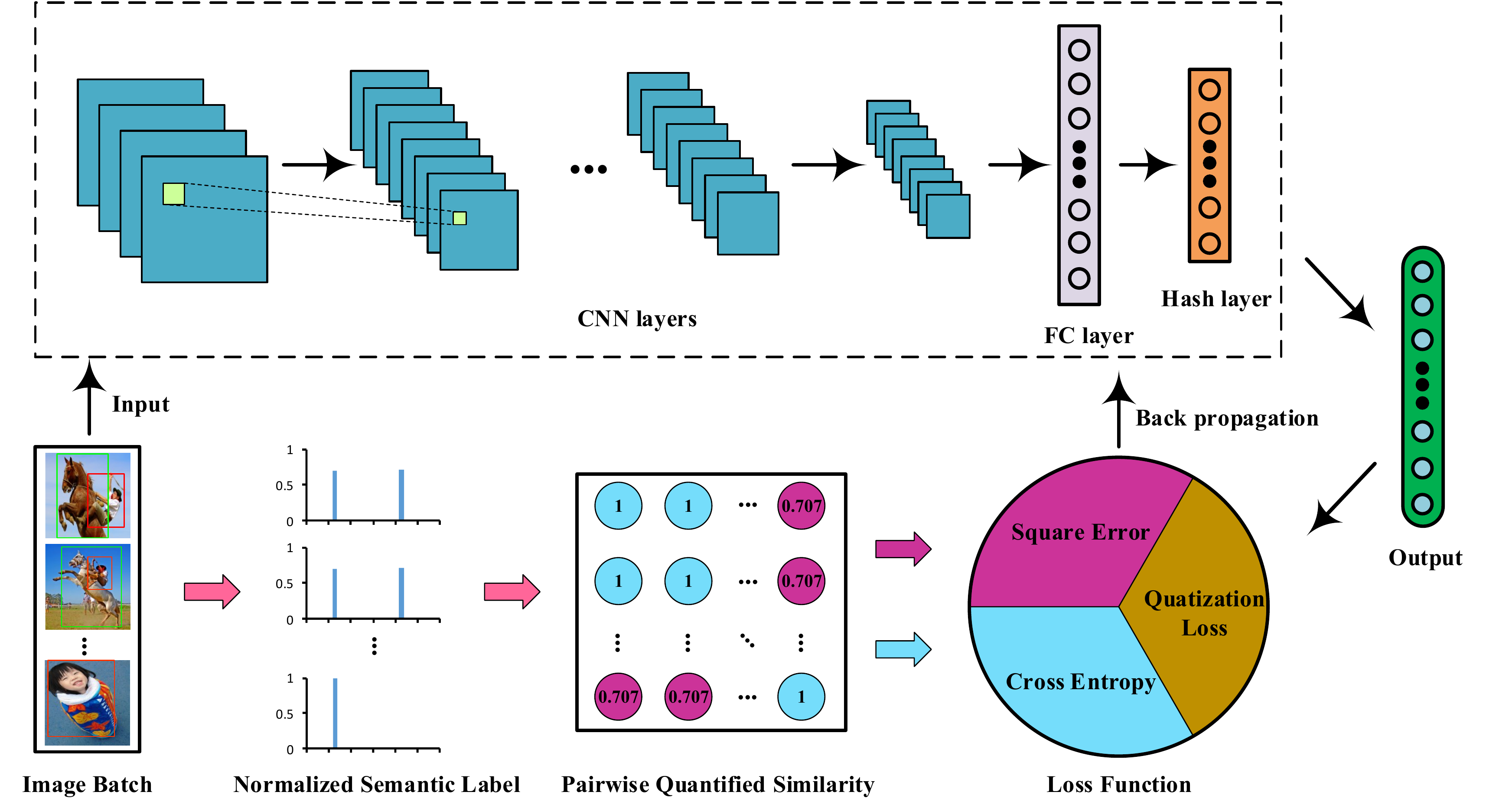}}
\caption{An overview of the proposed deep hashing learning method. The top frame shows the deep architecture of neural network that produces the hash codes.
The bottom frame shows the processing of pairwise quantified similarity and loss function construction. Cross entropy loss and mean square error loss are combined to preserving fine-grained pairwise similarity and a quantization loss is adapted to impose constraints for compact hash coding.
}
\label{fig:two}
\end{figure*}

Figure~\ref{fig:two} shows the pipeline of the proposed deep hashing network for supervised hash-code learning. The proposed method accepts input images in a pairwise form
($I_{i}$,$I_{j}$,$s_{ij}$) and processes them through the deep representation learning and hash coding.
It includes a sub-network with multiple convolution/pooling layers to perform image abstraction, fully-connected layer to approximate optimal dimension-reduced representation, and hash layer to generate ${q}$-bits hash codes. In this framework, a pairwise similarity loss is introduced for similarity-preserving learning, and a quantization loss is used to control the quality of hashing. The pairwise similarity loss consists of two parts -- the cross entropy loss and the square error loss. Details will be introduced in the following of this section.

\subsection{Deep Network Architecture}
Since many deep hashing methods~\cite{zhu2016deep,cao2016deep,cao2017hashnet,wu2017deep} have adapted AlexNet \cite{krizhevsky2012imagenet} as base network, without loss of generality, we also adopt the AlexNet as our base network. AlexNet comprises of five convolutional layers $conv1$ - $conv5$ and three fully connected layers $fc6$ - $fc8$.
After each hidden layer, a nonlinear mapping $z_{i}^{l}=a^{l}(W^{l}z_{i}^{l-1}+b^{l})$ is learned by the activation function $a^{l}$, where $z_{i}^{l}$ is the $l$-th layer feature representation for the original input, $W^{l}$ and $b^{l}$ are the weight and bias parameters of the $l$-th layer. We replace the $fc8$ layer of the softmax classifier in the original AlexNet with a new fully-connected hashing layer with $q$ hidden nodes, which converts the learned deep features into a low-dimensional hash codes. In order to realize hash encoding, we introduce an activation function $a^{l}(x)=\frac{x}{|x|+1} $ to map the output of $fc8$ to be within (-1,1).
Notice that, our method can be easily extended to other deep networks, such as GoogLeNet~\cite{szegedy2015going} and VGG19~\cite{simonyan2014very}, we will conduct experiments on these two classical networks to demonstrate the extensibility of our method.

\subsection{Hash-Code Learning}
For efficient nearest neighbor search, the semantic similarity of original images should be preserved in the Hamming space.
In the following, we will discuss our proposed hashing methods with reference to `hard similarity' and `soft similarity' situation, respectively.

\subsubsection{Hard Similarity}
In these situations, according to the quantized pairwise similarity calculated by Eq.~(\ref{eq:s}), the similarity of pairwise images $s_{ij}$ can only get value 0 or 1, which is identical to the similarity definition in previous deep pair hashing methods~\cite{zhu2016deep,liu2016deep}.
Given the hash codes $B$ of all images and the pairwise similarity relation $S_{h} =\{s_{ij}\}$, the conditional probability $p(s_{ij}|B)$ of $s_{ij}$ can be defined as follows:

\begin{equation}
\label{eq:condition_prob}
p(s_{ij}|B)=
\begin{cases}
 \sigma (\Omega _{ij}) , & \ s_{ij} = 1, \\
 1 - \sigma (\Omega _{ij}) , & \ s_{ij} = 0,  \\
\end{cases}
\end{equation}

where $\sigma (x) = \frac{1}{1+e^{-x}}$ is the sigmoid function, which we use to transform the Hamming distance into a kind of measure of similarity.
Previous works have shown that the inner product $\langle \cdot,\cdot \rangle$ is a good metric of the Hamming distance to quantify the pairwise similarity~\cite{zhu2016deep,cao2016deep}. In this work, we construct an inner product $\Omega _{ij} =  \langle b_{i},b_{j} \rangle =  b_{i}^{T}b_{j}$.

Here, we adapt negative log-likelihood as cost function to measure the pairwise similarity loss, as formulated by Eq.~(\ref{eq:l1}),
\begin{equation}
\label{eq:l1}
\begin{split}
\mathcal L_{1} & = - \sum_{s_{ij} \in S_{h}}^{} log(p(s_{ij}|B)) \\
&= - \sum_{s_{ij} \in S_{h}}^{} \big(s_{ij} log(\sigma(\Omega _{ij}))+(1 - s_{ij})log(1-\sigma (\Omega _{ij}))\big).
\end{split}
\end{equation}

Then, substituting the sigmoid function $\sigma(\Omega _{ij})$ with $\frac{1}{1+e^{-\Omega _{ij}}}$, we get
\begin{equation}
\label{eq:l2}
\mathcal L_{1} = \sum_{s_{ij} \in S_{h}}^{} \big(log(1+ e^{\Omega _{ij}})- s_{ij}\Omega _{ij}\big).
\end{equation}

\subsubsection{Soft Similarity}
In this situation, the pairwise similarities defined by Eq.~(\ref{eq:s}) are continuous value, we apply mean square error function to preserve the similarity of hash codes to fit the soft similarity. Thus, the pairwise similarity loss can be defined as:
\begin{equation}
\label{eq:l}
\mathcal L_{2} =  \sum_{s_{ij} \in S_{s}}^{} (\frac{\langle b_{i},b_{j} \rangle+ q}{2}- s_{ij}\cdot q)^2.
\end{equation}
As the inner product $\langle b_{i},b_{j} \rangle$ is within [-$q$, $q$], the value of $\frac{\langle b_{i},b_{j} \rangle+ q}{2}$ will be non-negative and be within $[0, q]$, which has a same value range as $s_{ij}\cdot q$.

Although the cross entropy loss can also be used to measure the similarity error in the soft similarity, the mean square error loss shows better performance when multi-label images have more complicated semantic relation and more shared labels. We will discuss this in the experiments.

\subsubsection{Joint Learning}
For simultaneous learning of these two cases and make an unified form, we use $M_{ij}$ to mark the two cases, where $M_{ij}$ = 1 denotes the `hard similarity' case, and $M_{ij}$ = 0 denotes `soft similarity' case. Hence, the pairwise similarity loss is rewritten as:
\begin{equation}
\label{eq:L}
\begin{split}
\mathcal L = \sum_{s_{ij} \in S}^{} & [  M_{ij}(log(1+ e^{\Omega _{ij}})- s_{ij}\Omega _{ij})+  \\
              & \gamma \cdot (1-M_{ij})(\frac{\langle b_{i},b_{j} \rangle+ q}{2}- s_{ij}\cdot q)^2 ],
 \end{split}
\end{equation}
where $\gamma$ is a weight parameter to make a tradeoff between the cross entropy loss and mean square error loss.

It is challenging to directly optimize Eq.~(\ref{eq:L}), because the binary constraint $b_{i} \in {\{-1,1\}^{q}}$ requires thresholding the network outputs, which may result in the vanishing-gradient problem in back propagation during the training procedure. Following previous works~\cite{wang2014hashing,liu2016deep,zhu2016deep}, we apply the continuous relaxation to solve this problem. We use the output of deep hashing network $u$ as a substitute for binary code $b$. $\Omega _{ij}$ is redefined as $\alpha u_{i}^{T}u_{j}$, where $\alpha$ is a positive hyper-parameter to control the constraint bandwidth.
Since the network output is not the binary codes, we use a pairwise quantization loss to encourage the network output to be close to standard binary codes.
The pairwise quantization loss is defined as
\begin{equation}
\label{eq:Q}
\mathcal Q =  \sum_{i,j \in N}^{} (\| | u_{i}|-1 \|_{1} + \| | u_{j}|-1 \|_{1}),
\end{equation}
where 1 is a vector of all ones, $\| \cdot \|_{1}$ is the L1-norm of the vector, $| \cdot |$ is the element-wise absolute value operation. By integrating the pairwise similarity loss and pairwise quantization loss, the final cost loss is defined as
\begin{equation}
\label{eq:C}
\mathcal C =  \mathcal L +\lambda \mathcal Q,
\end{equation}
where $\lambda$ is a weight coefficient for controlling the quantization loss.

%\vspace{3mm}
\subsection{Learning Algorithm}
During the training process, the standard back-propagation algorithm with mini-batch gradient descent method is used to optimize the pairwise loss function. By combining Eq.~(\ref{eq:L}) and Eq.~(\ref{eq:Q}), we rewrite the optimization objective function $\mathcal C$ as follows:

\begin{equation}
\label{eq:C1}
\begin{split}
\mathcal C = & \mathcal L +\lambda \mathcal  Q    \\
  = & \sum_{i,j \in N}^{} [ M_{ij}(log(1+ e^{\alpha u_{i}^{T}u_{j}})- \alpha \cdot s_{ij}\cdot u_{i}^{T}u_{j})  \\
              & +\gamma \cdot (1-M_{ij})(\frac{u_{i}^{T}u_{j} + q}{2}- s_{ij}\cdot q)^2  \\
              & +\lambda \cdot (\| | u_{i}|-1 \|_{1} + \| | u_{j}|-1 \|_{1})] .
 \end{split}
\end{equation}

In order to employ back propagation algorithm to optimize the network parameters, we need to compute the derivative of the objective function. The sub-gradients of Eq.~(\ref{eq:C1}) w.r.t. $u_{ik}$ ($k$-th unit of the network output $u_{i}$) can be written as:
\begin{equation}
\label{eq:grad_L}
\begin{split}
 \frac{\partial \mathcal L}{\partial u_{ik}} =  &  \alpha \cdot  M_{ij} \sum_{j \in N}^{}  (\sigma (\Omega _{ij})-s_{ij} ) \cdot u_{jk} + \\   \gamma   \cdot
& (1-M_{ij}) \sum_{j \in N}^{} (u_{i}^{T}u_{j} + q- 2 \cdot s_{ij}\cdot q)  \cdot u_{jk}, \\
\end{split}
\end{equation}
and
\begin{equation}
\label{eq:grad_Q}
\frac{\partial \mathcal Q}{\partial u_{ik }} =
\begin{cases}
 1 , &  -1 < u_{ik } < 0, \\
 -1 , &   otherwise,
\end{cases}
\end{equation}

The gradient of $u_{ik}$ w.r.t. $\hat{z}_{ik}^{l}$ (raw representations of hash layer before activation) can be calculated by
\begin{equation}
\label{eq:grad_u}
\frac{\partial u_{ik}}{\partial \hat{z}_{ik}^{l}} = sgn(\hat{z}_{ik}^{l}) \cdot \frac{1}{(|\hat{z}_{ik}^{l}|+1)^2},
\end{equation}
where $sgn(\cdot)$ is an element-wise sign function and $\hat{z}_{i}^{l} = W^{l}z_{i}^{l-1}+b^{l}$ is the output of the $l$-th layer before activation. The gradient of the network parameter $W^{l}$ is
\begin{equation}
\label{eq:grad}
\frac{\partial \mathcal C}{\partial W^{l}} = \sum{}{} ( \frac{\partial \mathcal L}{\partial u_{i}} + \lambda  \frac{\partial \mathcal Q}{\partial u_{i}}) \frac{\partial u_{i}}{\partial \hat{z}_{i}^{l}}
\cdot z_{i}^{l-1}.
\end{equation}

Since we have computed sub-gradients of the hash layer, the rest of the back-propagation procedure can be done in the standard manner. Note that, after the learning procedure, we have not obtained the corresponding binary codes of input images yet. The network only generates approximate hash codes that have values within (-1,1). To finally get the hash codes and evaluate the efficacy of the trained network, we need to treat the test query data as input and forward propagate the network to generate hash codes by using Eq.~(\ref{eq:sgn}),
\begin{equation}
\label{eq:sgn}
b_{ik} = sgn(u_{ik}).
\end{equation}

In this way, we can train the deep neural network in an end-to-end fashion, and any new input images can be encoded into binary codes by the trained deep hashing model. Ranking the distance of these binary hash codes in the Hamming space, we can obtain an efficient image retrieval.

%\vspace{3mm}

\section{Experiments and Results}
\label{sec:experiment}
\subsection{Datasets}

To verify the performance of the proposed method, we compare the proposed method with several baseline methods on four widely used benchmark datasets, i.e., {NUS-WIDE}, {Flickr}, {VOC2012} and {IAPRTC12}.

\textbf{NUS-WIDE} \cite{chua2009nus} is a dataset containing 269,648 public web images. It is a multi-label dataset in which each image is annotated with one or more class labels from a total of 81 classes. We follow the settings in~\cite{liu2011hashing,lai2015simultaneous} to use the subset of images associated with the 21 most frequent labels, where each label associates with at least 5,000 images, resulting in a total of 195,834 images. We resize the images of this subset to 227$\times$227.

\textbf{Flickr} \cite{huiskes2008mir} is a dataset containing 25,000 images collected from Flickr. Each image contains at least one of the 38 semantic labels. We resize the images to 227$\times$227.

\textbf{VOC2012} \cite{everingham2010pascal} is a widely used dataset for object detection and segmentation, which contains 17,125 images, and each image belongs to at least one of the 20 semantic labels. We resize the images to 227 $\times$ 227.

\textbf{IAPRTC12} \cite{escalante2010the} contains 20,000 images with segmentation masks. Each region of segmentation has been assigned with a label from a total of 276 pre-defined categories. We resize the images to 227 $\times$ 227.

\subsection{Implementation Details}
For NUS-WIDE, we randomly select 100 images per class to form a test query set of 2,100 images, and 500 images per class to form the training set. For Flickr, VOC2012 and IAPRTC12, we randomly select 1,000 images as the test query set, and 4,000 images as the train set. The remaining images in each of the four datasets are taken as query database.

\begin{figure}[!t]
  \centering
  \includegraphics[width=1.0\linewidth]{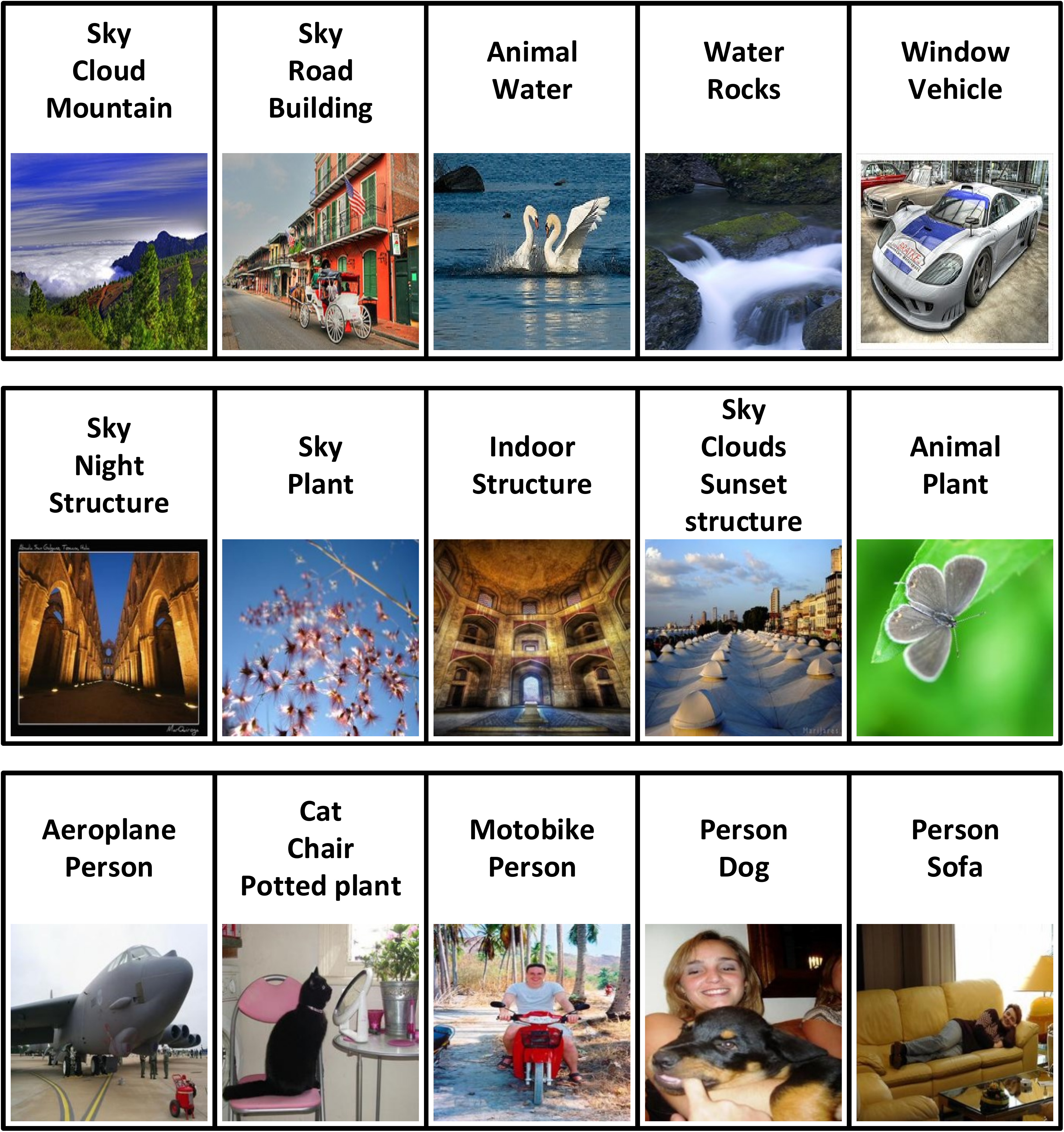}
  \caption{Some sample images. From top row to the bottom row are the samples from NUS-WIDE, Flickr, and VOC2012, respectively. The labels have been given for each image as provided by the datasets.}\label{fig:samples}
\end{figure}

We compare our method with several state-of-the-art hashing methods, including three unsupervised methods LSH~\cite{datar2004locality}, SH~\cite{weiss2009spectral} and ITQ~\cite{gong2013iterative}, and two traditional supervised methods MLH~\cite{norouzi2011minimal}, KSH~\cite{liu2012supervised}, and one classification-based deep hashing methods DLBHC~\cite{Lin2015Deep}, three deep hashing methods with coarse pairwise similarity definition--HashNet~\cite{cao2017hashnet}, DHN~\cite{zhu2016deep} and DQN~\cite{cao2016deep}, and one state-of-the-art deep hashing methods designed for multi-label image retrieval, DMSSPH~\cite{wu2017deep}. Notice that, although the Jaccard coefficient based similarity has been used as one multiplier of the weight for each training pair in HashNet~\cite{cao2017hashnet}, the similarity of pairwise image is still in a coarse way.
To make the comparison more extensive, four other typical deep hashing methods are also included, namely the DPSH~\cite{li2016feature}, DSRH~\cite{zhao2015deep}, DSDH~\cite{li2017deep} and DTSH~\cite{wang2016deep}. Based on a coarse pairwise similarity, DPSH constructs the cross-entropy loss and absolute quantization loss based on a coarse pairwise similarity, and DSDH directly learns the discrete hash codes with the auxiliary of classification mask. DTSH and DSRH are two triplet-based methods which learn hash function from the local triplet ranking relation~\cite{schroff2015facenet}.

We implement the proposed IDHN{\footnote{https://sites.google.com/site/qinzoucn/}} {\footnote{https://github.com/pectinid16/IDHN}}by the TensorFlow toolkit~\cite{abadi2016tensorflow}.
To make fair comparison, all deep hashing methods for comparison are reproduced by using the TensorFlow and based on the bone net of AlexNet. For other traditional methods, we use the open-source codes released by~\cite{sis}, which are implemented with MATLAB.
We fine-tune the convolutional layers $conv1$ - $conv5$ and fully-connected layers $fc6$ - $fc7$ with network weight parameters copied from the pre-trained model, and train the hashing layer $fc8$, all via back-propagation. We use the Adam method for stochastic optimization with a mini-batch size of 128, and the learning rate decay after each 500 iterations with a decay rate of 0.5.
For the deep learning based methods, we directly use the image pixels as the input. For the other traditional methods, i.e., LSH, SH, ITQ, MLH and KSH, feature maps of the fully-connected layer $fc7$ are extracted using the pre-trained model and taken as the input without other preprocessing.

\subsection{Metrics}
We evaluate the image retrieval quality using four widely-used metrics:
Average Cumulative Gains (ACG)~\cite{jarvelin2000ir}, Normalized Discounted Cumulative Gains (NDCG)~\cite{jarvelin2002cumulated}, Mean Average Precision (MAP)~\cite{baeza1999modern} and Weighted Mean Average Precision (WAP)~\cite{zhao2015deep}.

ACG represents the average number of shared labels between the query image and the top $n$ retrieved images. Given a query image $I_{q}$, the ACG score of the top $n$ retrieved images is calculated by
\begin{equation}
\label{eq:acg}
ACG@n = \frac{1}{n} \sum_{i}^{n} C(q,i),
\end{equation}
where $n$ denotes the number of top retrieval images and $C(q,i)$ is the number of shared class labels between $I_{q}$ and $I_{i}$.

NDCG is a popular evaluation metric in information retrieval. Given a query image $I_{q}$, the DCG score of top $n$ retrieved images is defined as
\begin{equation}
\label{eq:dcg}
DCG@n = \sum_{i}^{n} \frac{2^{C(q,i)}-1}{log(1 + i)}.
\end{equation}
Then, the normalized DCG (NDCG) score at the position $n$ can be calculated by $NDCG@n = \frac{DCG@n}{Z_{n}}$, where $Z_{n}$ is the maximum value of $DCG@n$, which constrains the value of NDCG in the range [0,1].

\begin{table}[!t]
  \centering
  %\normalsize
  \caption{Results of Mean Average Precision (MAP) for different parameter value of $\alpha$. Note that, $q$ denotes the length of the hash codes.}
  \begin{tabular}{c|c|c|c|c|c|c}
  \hlinew{1.2pt}
  \multirow{2}{*}{$\alpha$} &
  \multicolumn{2}{c|}{NUS-WIDE} &
  \multicolumn{2}{c|}{Flickr} &
  \multicolumn{2}{c}{VOC2012}\\
  \cline{2-3}
  \cline{4-5}
  \cline{6-7}
     & 24-bit & 48-bit & 24-bit & 48-bit & 24-bit & 48-bit\\
  \hline
    1/$q$ & 0.7141 & 0.7194 & 0.7878 & 0.8049 & 0.6185 & 0.6374 \\
  \hline
   5/$q$ & \textbf{0.7560} & \textbf{0.7681} & \textbf{0.8462} & \textbf{0.8515} & 0.6874 & 0.7032 \\
  \hline
   10/$q$ & 0.7498 & 0.7661 & 0.8425 & 0.8472 & \textbf{0.6886} & \textbf{0.7087} \\
  \hline
   20/$q$ & 0.7310 & 0.7528 & 0.8394 & 0.8492 & 0.6682 & 0.7014 \\
  \hline
   1 & 0.7202 & 0.7079 & 0.8367 & 0.8428 & 0.6644 & 0.6674 \\
  \hlinew{1.2pt}
  \end{tabular}
  \label{tb:alpha}
\end{table}

\begin{table}[!t]
  \centering
  %\normalsize
  \caption{Results of Mean Average Precision (MAP) for different parameter values of $\lambda$.}
  \begin{tabular}{c|c|c|c|c|c|c}
  \hlinew{1.2pt}
  \multirow{2}{*}{$\lambda$} &
  \multicolumn{2}{c|}{NUS-WIDE} &
  \multicolumn{2}{c|}{Flickr} &
  \multicolumn{2}{c}{VOC2012}\\
  \cline{2-3}
  \cline{4-5}
  \cline{6-7}
     & 24-bit & 48-bit & 24-bit & 48-bit & 24-bit & 48-bit\\
  \hline
    0 & 0.7259 & 0.7224 & 0.8020 & 0.7731 & 0.6393 & 0.6934 \\
  \hline
    0.01 & 0.7263 & 0.7241 & 0.7983 & 0.7719 & 0.6419 & 0.6890 \\
  \hline
    0.1 & \textbf{0.7345} & \textbf{0.7298} & \textbf{0.8052} & \textbf{0.7756} & \textbf{0.6428} & \textbf{0.6954} \\
  \hline
    1.0 & 0.6662 & 0.6743 & 0.7651 & 0.7628 & 0.6334 & 0.6525 \\
  \hline
    10.0 & 0.5382 & 0.5731 & 0.6878 & 0.6952 & 0.6228 & 0.4624 \\
  \hlinew{1.2pt}
  \end{tabular}
  \label{tb:lambda}
\end{table}

\begin{table}[!t]
  \centering
  %\normalsize
  \caption{Results of Mean Average Precision (MAP) for different parameter value of $\gamma$.}
  \begin{tabular}{c|c|c|c|c|c|c}
  \hlinew{1.2pt}
  \multirow{2}{*}{$\gamma$} &
  \multicolumn{2}{c|}{NUS-WIDE} &
  \multicolumn{2}{c|}{Flickr} &
  \multicolumn{2}{c}{VOC2012}\\
  \cline{2-3}
  \cline{4-5}
  \cline{6-7}
     & 24-bit & 48-bit & 24-bit & 48-bit & 24-bit & 48-bit\\
  \hline
    0 & 0.7227 & 0.7309 & 0.8398 & 0.8454 & 0.6582 & 0.6721 \\
  \hline
    0.01/$q$ & 0.7287 & 0.7542 & 0.8410 & 0.8501 & 0.6628 & 0.6779 \\
  \hline
    0.1/$q$ & \textbf{0.7600} & \textbf{0.7692} & \textbf{0.8462} & \textbf{0.8515} & \textbf{0.6874} & \textbf{0.7032} \\
  \hline
    1/$q$ & 0.7288 & 0.7275 & 0.8002 & 0.7757 & 0.6842 & 0.6933 \\
  \hline
    10/$q$ & 0.6693 & 0.6685 & 0.7179 & 0.7194 & 0.6640 & 0.6677 \\
  \hlinew{1.2pt}
  \end{tabular}
  \label{tb:gamma}
\end{table}

\begin{figure*}[!t]
  \centering

  \begin{minipage}{1.0\linewidth}
    \includegraphics[width=1.0\linewidth]{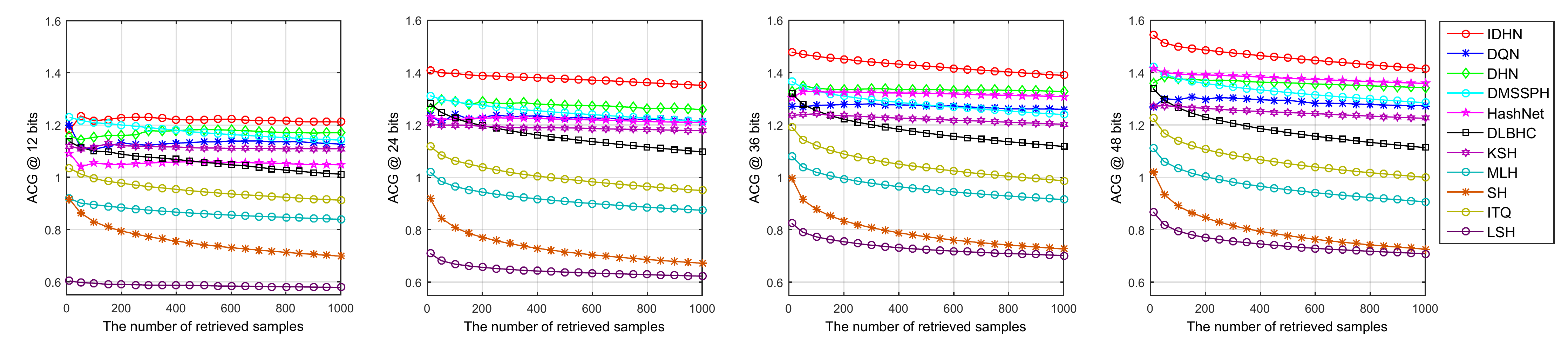}
  \end{minipage}

  \begin{minipage}{1.0\linewidth}
    \includegraphics[width=1.0\linewidth]{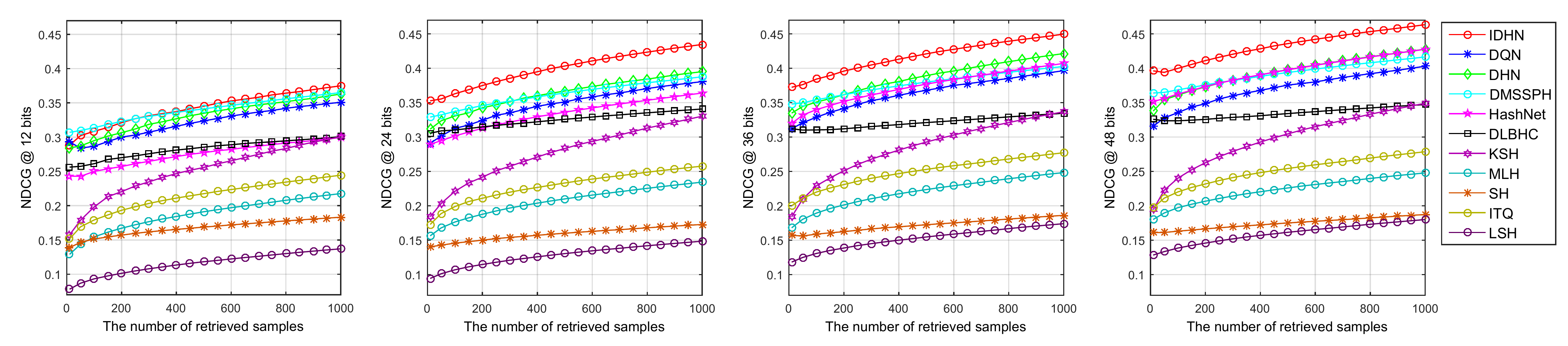}
  \end{minipage}

  \begin{minipage}{1.0\linewidth}
    \includegraphics[width=1.0\linewidth]{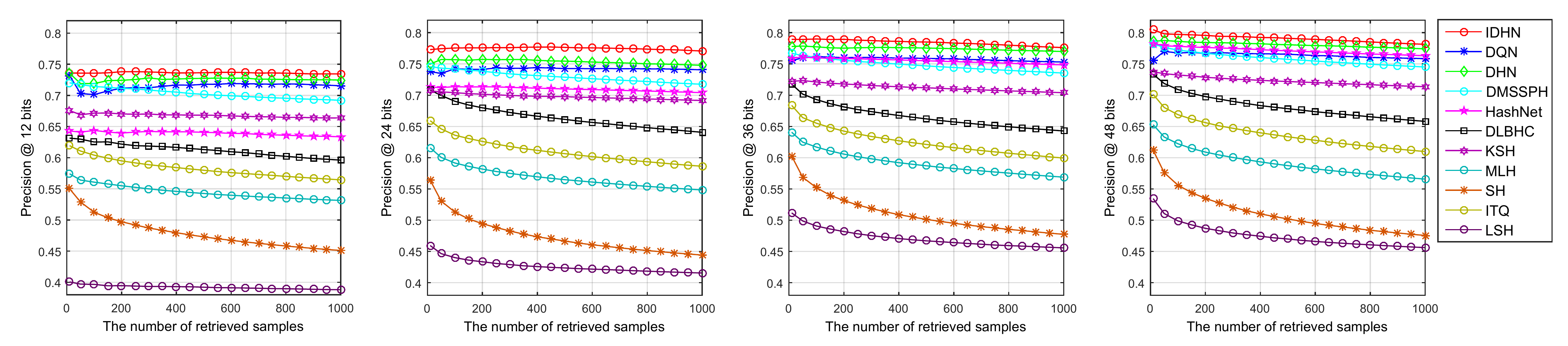}
  \end{minipage}
\caption{Performance of different methods on the NUS-WIDE dataset. From top to bottom, there are ACG, NDCG and precision curves w.r.t. different top returned samples with hash codes of 12, 24, 36 and 48 bits, respectively.}

\label{fig:NUS}
\end{figure*}

\begin{figure*}[!t]
  \centering

  \begin{minipage}{1.0\linewidth}
    \includegraphics[width=1.0\linewidth]{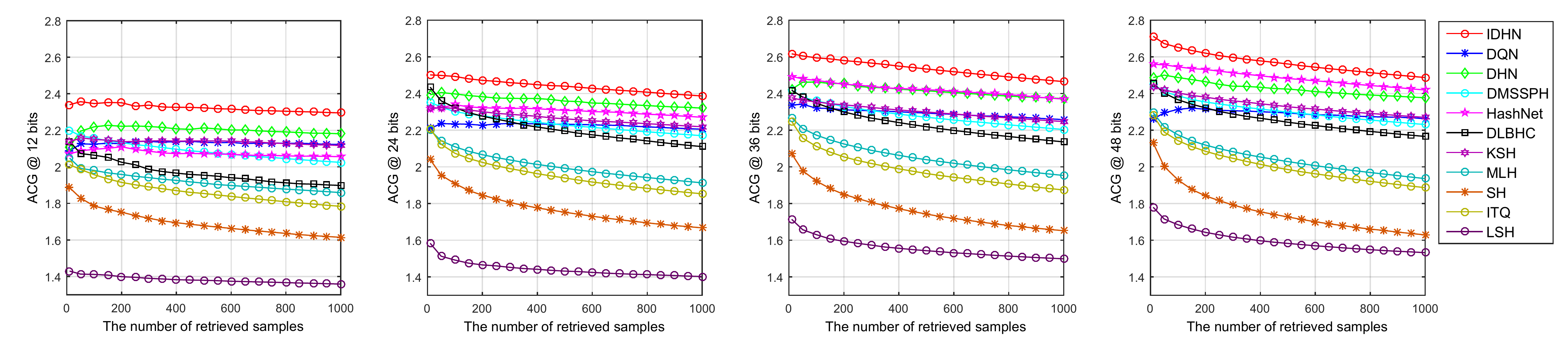}
  \end{minipage}

  \begin{minipage}{1.0\linewidth}
    \includegraphics[width=1.0\linewidth]{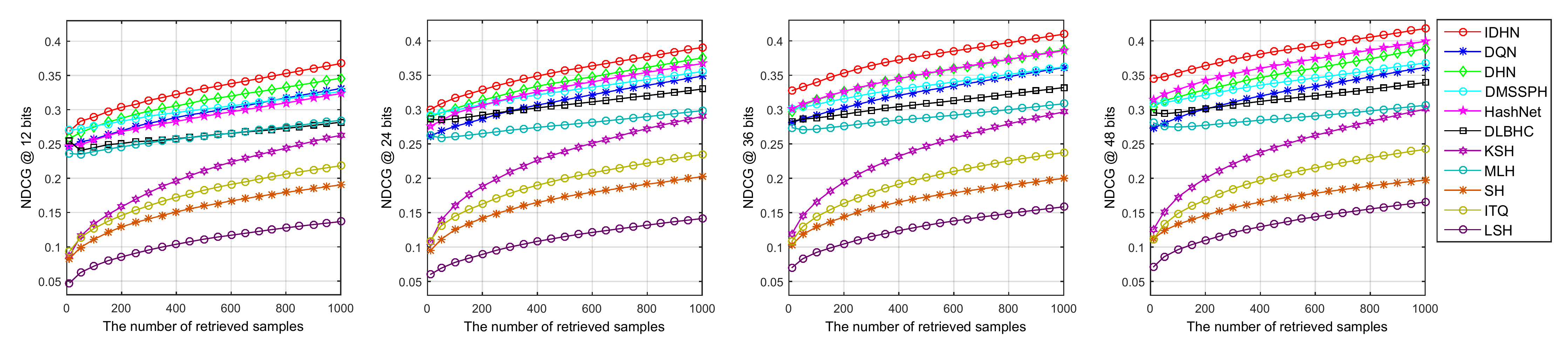}
    \end{minipage}

  \begin{minipage}{1.0\linewidth}
    \includegraphics[width=1.0\linewidth]{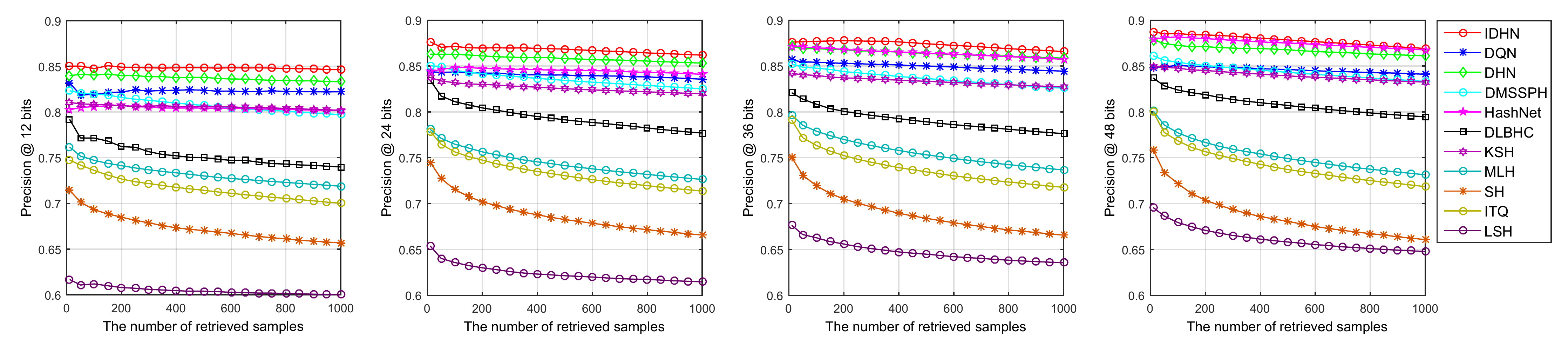}
  \end{minipage}
\caption{Performance of different methods on the Flickr dataset. From top to bottom, there are ACG, NDCG and precision curves w.r.t. different top returned samples with hash codes of 12, 24, 36 and 48 bits, respectively.}
\label{fig:flickr25k}
\end{figure*}

\begin{figure*}[!t]
  \centering

\begin{minipage}{1.0\linewidth}
    \includegraphics[width=1.0\linewidth]{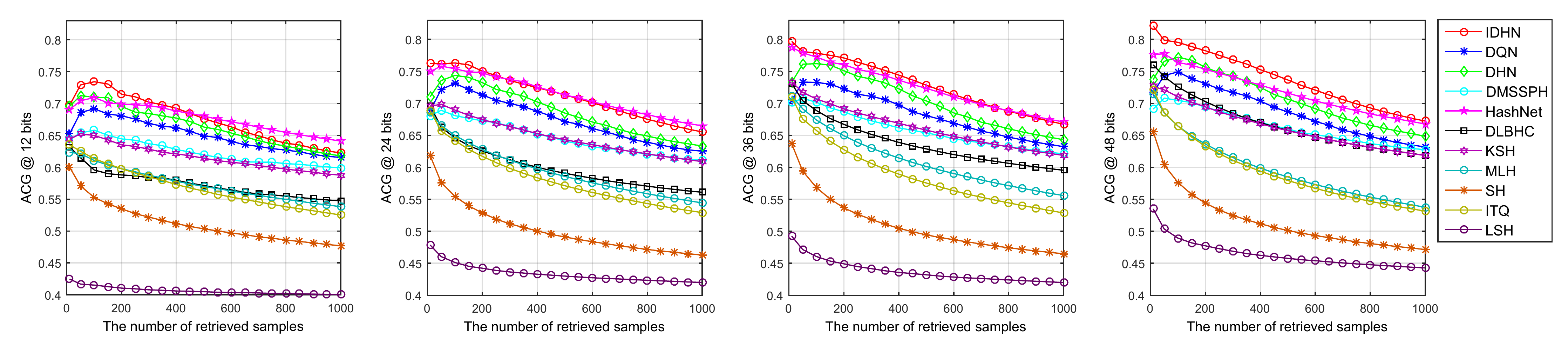}
\end{minipage}

\begin{minipage}{1.0\linewidth}
    \includegraphics[width=1.0\linewidth]{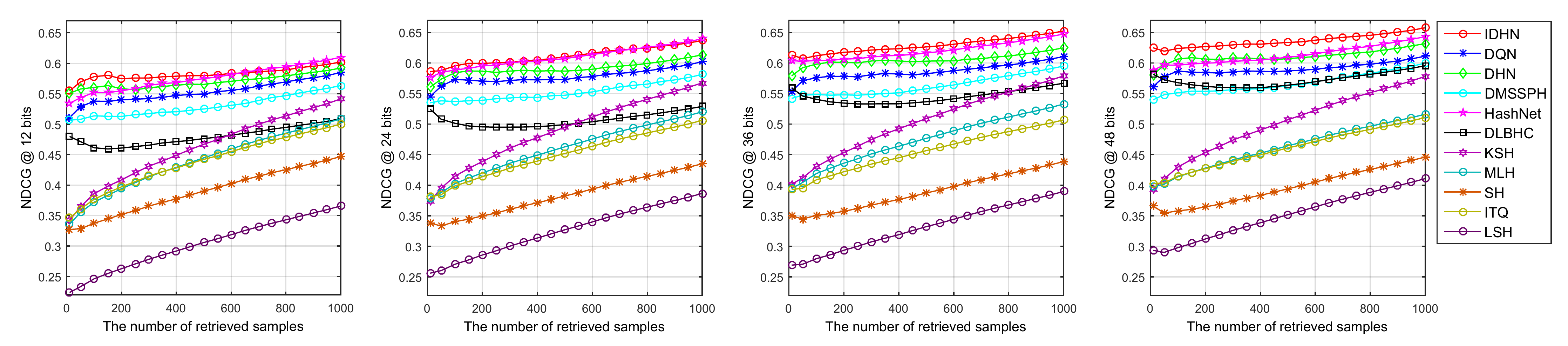}
\end{minipage}

  \begin{minipage}{1.0\linewidth}
    \includegraphics[width=1.0\linewidth]{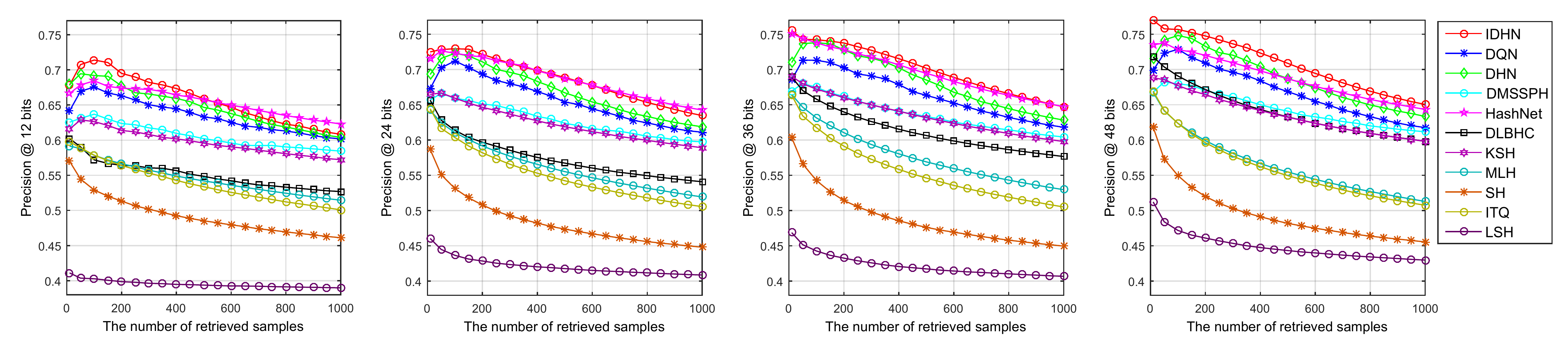}
  \end{minipage}

\caption{Performance of difference methods on the VOC2012 dataset. From top to bottom, there are ACG, NDCG and precision curves w.r.t. different number of top returned samples with hash codes of 12, 24, 36 and 48 bits, respectively.}
\label{fig:voc2012}
\end{figure*}

\begin{figure*}[!t]
  \centering
  \centerline{\includegraphics[width=1\linewidth]{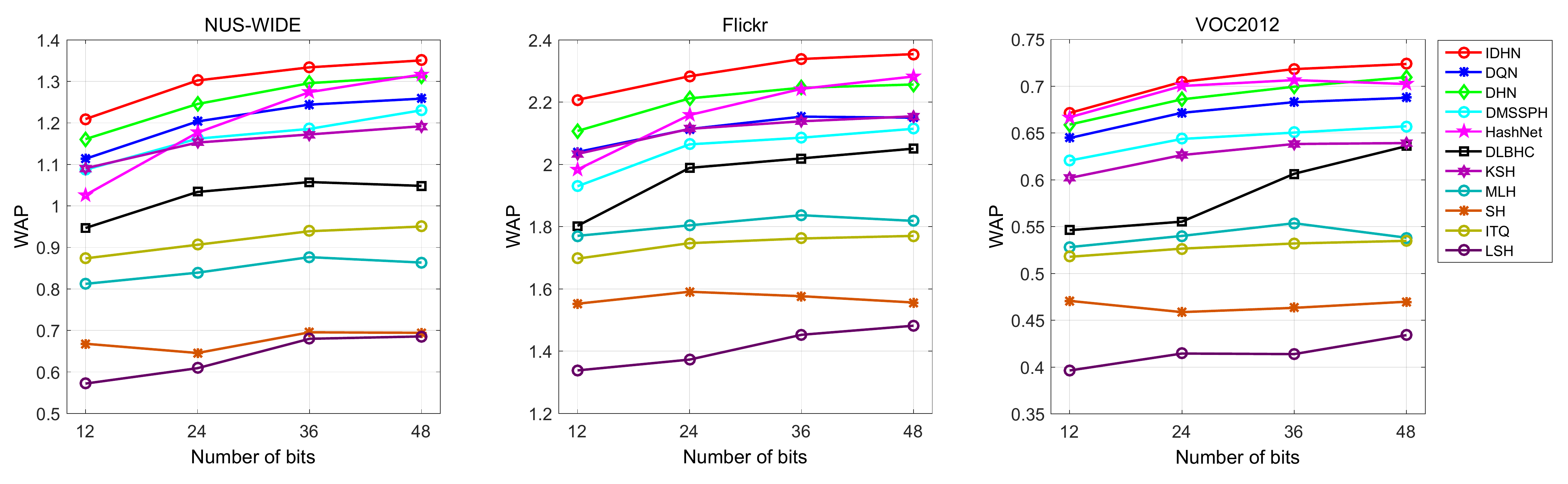}}
\caption{Comparison of the IDHN method and other compared methods on $WAP@5000$ results.}
\label{fig:wap}
\end{figure*}

MAP is the mean of average precision for each query, which can be calculated by
\begin{equation}
\label{eq:map}
MAP = \frac{1}{Q} \sum_{q}^{Q} AP(q),
\end{equation}
where
\begin{equation}
\label{eq:ap}
AP(q) =\frac{1}{N_{Tr}(q)@n} \sum_{i}^{n} \Big(Tr(q,i)  \frac{N_{Tr}(q)@i}{i}\Big),
\end{equation}
and $Tr(q,i) \in \{0,1\} $ is an indicator function that if $I_{q}$ and $I_{i}$ share some class labels, $Tr(q,i)$ = 1; otherwise $Tr(q,i)$ = 0. $Q$ is the numbers of query sets and $N_{Tr}(q)@i$ indicates the number of relevant images w.r.t. the query image $I_{q}$ within the top $i$ images.

The definition of WAP is similar with MAP. The only difference is that WAP computes the average ACG scores at each top $n$ retrieved image rather than average precision. WAP can be calculated by
\begin{equation}
\label{eq:wap}
WAP =\frac{1}{Q} \sum_{q}^{Q} \Big(\frac{1}{N_{Tr}(q)@n} \sum_{i}^{n} \big(Tr(q,i) \times ACG@i\big)\Big).
\end{equation}

\subsection{Results}
\subsubsection{Parameter analysis}
The parameter $\alpha$ is used to control the range of inner product value after normalization. We notice that the gradient of large absolute value is very small in the sigmoid function, which may cause gradient vanishing. In order to avoid this and accelerate the convergence, we employ the parameter $\alpha$ and set its value according to the length $q$ of the hash codes.
Table~\ref{tb:alpha} shows the results of the proposed method by using different values of $\alpha$. We set $\alpha$ = $\frac{5}{q}$ to constrain the result of $\Omega_{ij}$ to be within [-5,5], which is relatively a suitable range. $\gamma$ and $\lambda$ are the coefficient of mean-square-error loss and quantization loss, respectively.
We first study the influence of $\lambda$. It can be seen from Table~\ref{tb:lambda}, it achieves the best results at $\lambda$=0.1.
It is because that a larger $\lambda$ will result in more discrete but less similarity-preserved hash codes, and a smaller $\lambda$ will make the quantization loss less effective.
We also examine the influence of $\gamma$. It can be seen from Table~\ref{tb:gamma}, $\gamma$=$\frac{0.1}{q}$ leads to the highest performance among all settings. By dividing $q$, the gradient of mean square error loss in Eq.~(\ref{eq:grad_L}) can be adaptively adjusted within a suitable range. Too larger or too smaller a $\gamma$ value will destroy the balance between the cross-entropy loss and the mean-square-error loss.

\begin{table}[!t]
  \centering
  %\normalsize
  \caption{Results of Mean Average Precision (MAP) for different numbers of bits on NUS-WIDE dataset.}
  \begin{tabular}{c|c|c|c|c}
  \hlinew{1.2pt}
    Methods & 12-bit & 24-bit & 36-bit & 48-bit  \\
  \hline
    \textbf{IDHN} & \textbf{0.7292} & \textbf{0.7585} & \textbf{0.7639} & \textbf{0.7692} \\
    DQN\cite{cao2016deep} & 0.7106 & 0.7327 & 0.7454 & 0.7493 \\
    DHN\cite{zhu2016deep} & 0.7187 & 0.7399 & 0.7595 & 0.7637 \\
    DMSSPH\cite{wu2017deep} & 0.6713 & 0.6993 & 0.7173 & 0.7273 \\
    HashNet\cite{cao2017hashnet} & 0.6429 & 0.6938 & 0.7371 & 0.7501 \\
    DLBHC\cite{Lin2015Deep} & 0.5696 & 0.6160 & 0.6214 & 0.6351 \\
    KSH\cite{liu2012supervised} & 0.6556 & 0.6825 & 0.6934 & 0.7024 \\
    MLH\cite{norouzi2011minimal} & 0.5184 & 0.5319 & 0.5512 & 0.5458 \\
    SH\cite{weiss2009spectral} & 0.4368 & 0.4412 & 0.4616 & 0.4596 \\
    ITQ\cite{gong2013iterative} & 0.5469 & 0.5666 & 0.5785 & 0.5876 \\
    LSH\cite{datar2004locality} &0.3854 & 0.4085 & 0.4452 & 0.4453 \\
  \hlinew{1.2pt}
  \end{tabular}
  \label{tb:one}
\end{table}

\begin{table}[!t]
  \centering
  %\normalsize
  \caption{Results of Mean Average Precision (MAP) for different numbers of bits on Flickr dataset.}
  \begin{tabular}{c|c|c|c|c}
  \hlinew{1.2pt}
    Methods & 12-bit & 24-bit & 36-bit & 48-bit  \\
  \hline
    \textbf{IDHN} & \textbf{0.8327} & \textbf{0.8469} & \textbf{0.8490} & \textbf{0.8515} \\
    DQN\cite{cao2016deep} & 0.8092 & 0.8227 & 0.8298 & 0.8270 \\
    DHN\cite{zhu2016deep} & 0.8227 & 0.8393 & 0.8446 & 0.8471 \\
    DMSSPH\cite{wu2017deep} & 0.7800 & 0.8080 & 0.8096 & 0.8159 \\
    HashNet\cite{cao2017hashnet} & 0.7909 & 0.8262 & 0.8414 & 0.8483  \\
    DLBHC\cite{Lin2015Deep} & 0.7236 & 0.7566 & 0.7573 & 0.7761 \\
    KSH\cite{liu2012supervised} & 0.7907 & 0.8070 & 0.8141 & 0.8181 \\
    MLH\cite{norouzi2011minimal} & 0.7033 & 0.7073 & 0.7163 & 0.7103 \\
    SH\cite{weiss2009spectral} & 0.6451 & 0.6512 & 0.6505 & 0.6463 \\
    ITQ\cite{gong2013iterative} & 0.6845 & 0.6950 & 0.6973 & 0.6978 \\
    LSH\cite{datar2004locality} & 0.5968 & 0.6086 & 0.6265 & 0.6369 \\
  \hlinew{1.2pt}
  \end{tabular}
  \label{tb:two}
\end{table}

\begin{table}[!t]
  \centering
  %\normalsize
  \caption{Results of Mean Average Precision (MAP) for different numbers of bits on VOC2012 dataset.}
  \begin{tabular}{c|c|c|c|c}
  \hlinew{1.2pt}
    Methods & 12-bit & 24-bit & 36-bit & 48-bit  \\
  \hline
    \textbf{IDHN} & \textbf{0.6561} & \textbf{0.6874} & \textbf{0.6991} & \textbf{0.7032} \\
    DQN\cite{cao2016deep} & 0.6303 & 0.6564 & 0.6675 & 0.6716 \\
    DHN\cite{zhu2016deep} & 0.6445 & 0.6704 & 0.6829 & 0.6928 \\
    DMSSPH\cite{wu2017deep} & 0.6064 & 0.6298 & 0.6343 & 0.6420 \\
    HashNet\cite{cao2017hashnet} & 0.6502 & 0.6809 & 0.6856 & 0.6871  \\
    DLBHC\cite{Lin2015Deep} & 0.5284 & 0.5372 & 0.5895 & 0.6173 \\
    KSH\cite{liu2012supervised} & 0.5874 & 0.6088 & 0.6196 & 0.6209 \\
    MLH\cite{norouzi2011minimal} & 0.5074 & 0.5179 & 0.5305 & 0.5263 \\
    SH\cite{weiss2009spectral} & 0.4465 & 0.4453 & 0.4493 & 0.4552 \\
    ITQ\cite{gong2013iterative} & 0.4966 & 0.5054 & 0.5110 & 0.5134 \\
    LSH\cite{datar2004locality} & 0.3866 & 0.4039 & 0.4118 & 0.4222 \\
  \hlinew{1.2pt}
  \end{tabular}
  \label{tb:three}
\end{table}
\subsubsection{Comparison with state-of-the-art methods}
The results of MAP on NUS-WIDE, Flickr and VOC2012 datasets are shown from Table~\ref{tb:one} to \ref{tb:three}. It can be observed that, the proposed IDHN method substantially outperforms all the comparison methods on these three datasets.
Among the traditional hashing methods, KSH obtains the best results. Compared with KSH, the proposed method IDHN achieves an improvement of about 7.2\%, 3.8\% and 7.7\% in average MAP for different bits on NUS-WIDE, Flickr and VOC2012, respectively.
It can also be observed from Table \ref{tb:one} to \ref{tb:three}, the deep learning methods have obtained largely improved performance over the three traditional methods. Compared with DMSSPH, a deep hashing method designed for multi-label image retrieval, the proposed IDHN achieves an improvement of about 5.0\%, 4.1\% and 5.7\% in average MAP on the three datasets, respectively. For other three deep hashing methods with coarse similarity definition, i.e., DQN, DHN and HashNet, high MAP values have been obtained. However, compared with these three methods, IDHN still holds a higher average MAP with significance, which is 2.0\%, 2.2\%, 2.9\% higher than DQN, 0.9\%, 0.5\%, 1.3\% higher than DHN, and 4.8\%, 1.8\%, 1.0\% higher than HashNet on the three datasets, respectively. It indicates the effectiveness and the advantage of proposed fine pairwise similarity, which can preserve more fine-grained semantic similarity than the coarse similarity.

We also evaluate the performance of these methods on IAPRTC12. The results in Table~\ref{tb:iaprtc} show that, the proposed method achieves the best performance among all methods. The number of labels in IAPRTC12 is 276, which is much larger than NUS-WIDE's 21, Flickr's 38 and VOC2012's 20. The above results simply indicate that the proposed method can be effective in handling multi-label images with a small or large number of labels.

Figure~\ref{fig:NUS} shows the ACG, NDCG, and precision curves of compared hashing methods w.r.t. different numbers of top returned images with 12, 24, 36 and 48 bits on NUS-WIDE, respectively. On these metrics, the advantage of the proposed method is not great on 12 bits compared to deep hashing methods, DHN and DMSSPH. It may be because that a shorter code is less effective in representing the semantic similarity of multi-label images in a large-scale dataset. When the code length increases, the performance of the proposed IDHN improves rapidly and shows obvious advantage than other compared methods. DLBHC shows the worse results among these deep hashing methods, since it is essentially a classification task using the class label as supervised information rather than retrieval task to preserve the semantic similarity.

\begin{table}[!t]
  \centering
  %\normalsize
  \caption{Results of Mean Average Precision (MAP) for different numbers of bits on IAPRTC-12 dataset.}
  \begin{tabular}{c|c|c|c|c}
  \hlinew{1.2pt}
    Methods & 12-bit & 24-bit & 36-bit & 48-bit  \\
  \hline
    \textbf{IDHN} & \textbf{0.5495} & \textbf{0.5697} & \textbf{0.5779} & \textbf{0.5859} \\
    DQN\cite{cao2016deep} & 0.5409 & 0.5672 & 0.5728 & 0.5774 \\
    DHN\cite{zhu2016deep} &  0.5412 & 0.5672 & 0.5762 & 0.5781 \\
    DMSSPH\cite{wu2017deep} & 0.4412 & 0.4745 & 0.4877 & 0.4928 \\
    HashNet\cite{cao2017hashnet} & 0.4912 & 0.5242 & 0.5472 & 0.5612  \\
    DLBHC\cite{Lin2015Deep} & 0.3218 & 0.3977 & 0.4199 & 0.4418 \\
    KSH\cite{liu2012supervised} & 0.5004 & 0.5211 & 0.5313 & 0.5363 \\
    MLH\cite{norouzi2011minimal} & 0.4618 & 0.4725 & 0.4763 & 0.4791 \\
    SH\cite{weiss2009spectral} & 0.3883 & 0.3793 & 0.3930 & 0.3905 \\
    ITQ\cite{gong2013iterative} & 0.4507 & 0.4628 & 0.4659 & 0.4700 \\
    LSH\cite{datar2004locality} & 0.3497 & 0.3553 & 0.3712 & 0.3891 \\
  \hlinew{1.2pt}
  \end{tabular}
  \label{tb:iaprtc}
\end{table}

\begin{table*}[htbp]
%\scriptsize
%\footnotesize
%\small
\centering
\caption{Results of MAP at different numbers of bits on four datasets.}\vspace{-2mm}

\begin{tabular}{|c|c|c|c|c|c|c|c|c|c|c|c|c|c|c|}
\hline
  \multirow{2}{*}{Methods} &
  \multicolumn{3}{c|}{NUS-WIDE} &
  \multicolumn{3}{c|}{Flickr} &
  \multicolumn{3}{c|}{VOC2012}&
  \multicolumn{3}{c|}{IAPRTC12}\\
  \cline{2-4}
  \cline{5-7}
  \cline{8-10}
  \cline{11-13}
    & 12-bit & 24-bit  & 48-bit & 12-bit & 24-bit  & 48-bit & 12-bit & 24-bit  & 48-bit & 12-bit & 24-bit & 48-bit \\
%    \hline \hline
%      \multicolumn{13}{|c|}{MAP}  \\
    \hline
  \multirow{1}{*}{IDHN} & \textbf{0.7296} & \textbf{0.7586} &  \textbf{0.7692} & \textbf{0.8327} & \textbf{0.8469} &  \textbf{0.8515} &  \textbf{0.6561} &  \textbf{0.6845} &  \textbf{0.7032} & \textbf{0.5495} & \textbf{0.5697} & \textbf{0.5859} \\
  \multirow{1}{*}{DPSH~\cite{li2016feature}} & 0.7088 & 0.7461 & 0.7606 & 0.8150 & 0.8369 & 0.8467 & 0.6123 & 0.6299 & 0.6461 & 0.5442 & 0.5548 & 0.5607 \\
  \multirow{1}{*}{DSDH~\cite{li2017deep}} & 0.6333 & 0.6918 & 0.7317 & 0.7637 & 0.7824 & 0.8195 & 0.5913 & 0.6343 & 0.6494 & 0.5120 & 0.5458 & 0.5620 \\
  %\hline
  \multirow{1}{*}{DSRH~\cite{zhao2015deep}} & 0.7029 & 0.7156 & 0.7321 & 0.8002 & 0.8140 & 0.8238 & 0.6518 & 0.6618 & 0.6675 & 0.5265 & 0.5397 & 0.5475 \\
  %\hline
  \multirow{1}{*}{DTSH~\cite{wang2016deep}} & 0.6972 & 0.7132 & 0.7462 & 0.8174 & 0.8283 & 0.8413 & 0.6552 & 0.6749 & 0.6795 & 0.5274 & 0.5452 & 0.5539 \\
  \hline
 \end{tabular}
\label{table:compare}
\end{table*}

\begin{table*}[htbp]
%\scriptsize
%\tiny
%\small
\centering
\caption{Results of MAP, WAP, ACG and NDCG at different numbers of bits on NUS-WIDE, Flickr and VOC2012 datasets.}\vspace{-2mm}
{
\begin{tabular}{|c|c|c|c|c|c|c|c|c|c|c|c|c|}
\hline
  \multirow{2}{*}{Methods} &
  \multicolumn{4}{c|}{NUS-WIDE} &
  \multicolumn{4}{c|}{Flickr} &
  \multicolumn{4}{c|}{VOC2012}\\
  \cline{2-5}
  \cline{6-9}
  \cline{10-13}
    & 12-bit & 24-bit & 36-bit & 48-bit & 12-bit & 24-bit & 36-bit & 48-bit & 12-bit & 24-bit & 36-bit & 48-bit\\
    \hline \hline
      \multicolumn{13}{|c|}{MAP}  \\
    \hline
  \multirow{1}{*}{IDHN} & \underline{0.7296} & \underline{0.7586} & \underline{0.7639} & \underline{0.7692} & \underline{0.8327} & \underline{0.8469} & \underline{0.8490} & \underline{0.8515} &  0.6561 & 0.6874 & 0.6991 & 0.7032 \\
  \multirow{1}{*}{IDHN-fine-ce} & 0.7237 & 0.7523 & 0.7601 & 0.7639 & 0.8252 & 0.8323 & 0.8406 & 0.8419 & \underline{0.6603} & \underline{0.6867} & \underline{0.7022} & \underline{0.7064}  \\
  %\hline
  \multirow{1}{*}{IDHN-fine-mse} & 0.7301 & 0.7553 & 0.7604 & 0.7641 & 0.8284 & 0.8373 & 0.8400 & 0.8439  & 0.6450 & 0.6712 & 0.6777 & 0.6803  \\
  %\hline
  \multirow{1}{*}{IDHN-coarse-ce} & 0.7203 & 0.7413 & 0.7469 & 0.7598 & 0.8234 & 0.8342 & 0.8360 & 0.8401 & 0.6329 & 0.6630 & 0.6699 & 0.6709  \\
  %\hline
  \multirow{1}{*}{IDHN-coarse-mse} & 0.6801 & 0.6921 & 0.7086 & 0.7116 & 0.7975 & 0.8045 & 0.8360 & 0.8401 & 0.6097 & 0.6359 & 0.6376 & 0.6398 \\
  %\hline
  \multirow{1}{*}{IDHN-GoogLeNet} & 0.7512 & 0.7696 & 0.7738 & 0.7796 & 0.8552 & 0.8653 & 0.8668 & 0.8697 & 0.7499 & 0.7741 & 0.7841 & 0.7898 \\
  %\hline
  \multirow{1}{*}{IDHN-VGG19} & \textbf{0.7638} & \textbf{0.7769} & \textbf{0.7851} & \textbf{0.7884} & \textbf{0.8746} & \textbf{0.8830} &\textbf{0.8834} & \textbf{0.8843} & \textbf{0.7706} &\textbf{ 0.7842} & \textbf{0.7926} & \textbf{0.8020} \\
  \hline \hline

  \multicolumn{13}{|c|}{WAP}  \\
  \hline
  \multirow{1}{*}{IDHN} & 2.2076 & \underline{2.2832} & \underline{2.3386} & \underline{2.3509} & 1.2092 & \underline{1.3022} & \underline{1.3339} & \underline{1.3506} & 0.6714 & 0.7048 & 0.7182 & 0.7237 \\
  \multirow{1}{*}{IDHN-fine-ce} & 2.2085 & 2.2662 & 2.3049 & 2.3204 & \underline{1.2573} & 1.2906 & 1.3120 & 1.3261 & \underline{0.6783} & \underline{0.7061} & \underline{0.7223} & \underline{0.7267}  \\
  %\hline
  \multirow{1}{*}{IDHN-fine-mse} & \underline{2.2141} & 2.2666 & 2.2845 & 2.3038 & 1.2467 & 1.3052 & 1.3248 & 1.3342  & 0.6612 & 0.6882 & 0.6915 & 0.6981  \\
  %\hline
  \multirow{1}{*}{IDHN-coarse-ce} & 2.1426 & 2.1918 & 2.2134 & 2.2317 & 1.1822 & 1.2348 & 1.2628 & 1.2954 & 0.6473 & 0.6787 & 0.6859 & 0.6869  \\
  %\hline
  \multirow{1}{*}{IDHN-coarse-mse} & 2.0046 & 2.0500 & 2.0592 & 2.0516 & 1.0523 & 1.0769 & 1.1198 & 1.1231 & 0.6235 & 0.6507 & 0.6521 & 0.6545 \\
  %\hline
  \multirow{1}{*}{IDHN-GoogLeNet} & 2.2922 & 2.3302 & 2.3771 & 2.3879 & 1.2687 & \textbf{1.3212} & 1.3364 & \textbf{1.3524} & 0.7699 & 0.7956 & 0.8061 & 0.8119 \\
  %\hline
  \multirow{1}{*}{IDHN-VGG19} & \textbf{2.3422} & \textbf{2.4027} & \textbf{2.4357} & \textbf{2.4362} & \textbf{1.2882} & 1.3048 & \textbf{1.3405} & 1.3430 & \textbf{0.7907} & \textbf{0.8066} & \textbf{0.8160} & \textbf{0.8266} \\
  \hline \hline

%  \multirow{1}{*}{} &
  \multicolumn{13}{|c|}{ACG}  \\
  \hline
  \multirow{1}{*}{IDHN} & \underline{2.0376} & \underline{2.0775} & \underline{2.0977} & \underline{2.0919} & 1.1560 & 1.1942 & \underline{1.2110} & \underline{1.2135} & 0.5426 & \underline{0.5555} & \underline{0.5583} & \underline{0.5586} \\
  \multirow{1}{*}{IDHN-fine-ce} & 1.9838 & 2.0185 & 2.0362 & 2.0441 & 1.1498 & 1.1625 & 1.1750 & 1.1824 & \underline{0.5483} & 0.5516 & 0.5563 & 0.5546  \\
  %\hline
  \multirow{1}{*}{IDHN-fine-mse} & 2.0260 & 2.0503 & 2.0630 & 2.0680 & \underline{1.1634} & \underline{1.2003} & 1.2100 & 1.2118  & 0.5468 & 0.5507 & 0.5503 & 0.5517  \\
  %\hline
  \multirow{1}{*}{IDHN-coarse-ce} & 1.9731 & 1.9859 & 2.0009 & 2.0033 & 1.1444 & 1.1770 & 1.1902 & 1.2035 & 0.5362 & 0.5432 & 0.5491 & 0.5501  \\
  %\hline
  \multirow{1}{*}{IDHN-coarse-mse} & 1.8735 & 1.8821 & 1.8836 & 1.8861 & 1.0202 & 1.0372 & 1.0753 & 1.0798 & 0.5251 & 0.5312 & 0.5321 & 0.5334 \\
  %\hline
  \multirow{1}{*}{IDHN-GoogLeNet} & 2.1277 & 2.1401 & 2.1827 & 2.1856 & \textbf{1.2021} & \textbf{1.2077} & 1.2163 & 1.2269 & 0.5702 & 0.5701 & 0.5733 & 0.5738 \\
  %\hline
  \multirow{1}{*}{IDHN-VGG19} & \textbf{2.1575} & \textbf{2.2227} & \textbf{2.2436} &\textbf{ 2.2458} & 1.1988 & 1.2075 & \textbf{1.2268} & \textbf{1.2282} & \textbf{0.5740} & \textbf{0.5785} & \textbf{0.5767} & \textbf{0.5816} \\
  \hline \hline

  \multicolumn{13}{|c|}{NDCG}  \\
  \hline
  \multirow{1}{*}{IDHN} & \underline{0.5542} & \underline{0.5705} & \underline{0.5795} & \underline{0.5786} & 0.5240 & \underline{0.5602} & \underline{0.5716} & \underline{0.5780} & 0.7218 & 0.7502 & 0.7604 & 0.7641 \\
  \multirow{1}{*}{IDHN-fine-ce} & 0.5440 & 0.5591 & 0.5672 & 0.5699 & \underline{0.5407} & 0.5545 & 0.5623 & 0.5678 & \underline{0.7445} & \underline{0.7598} & \underline{0.7674} & \underline{0.7667}  \\
  %\hline
  \multirow{1}{*}{IDHN-fine-mse} & 0.5504 & 0.5621 & 0.5662 & 0.5705 & 0.5339 & 0.5593 & 0.5663 & 0.5687  & 0.7233 & 0.7355 & 0.7393 & 0.7415  \\
  %\hline
  \multirow{1}{*}{IDHN-coarse-ce} & 0.5408 & 0.5484 & 0.5531 & 0.5559 & 0.5102 & 0.5330 & 0.5440 & 0.5581 & 0.7006 & 0.7207 & 0.7257 & 0.7286  \\
  %\hline
  \multirow{1}{*}{IDHN-coarse-mse} & 0.5046 & 0.5095 & 0.5101 & 0.5108 & 0.4429 & 0.4527 & 0.4754 & 0.4795 & 0.6830 & 0.6855 & 0.6976 & 0.6989 \\
  %\hline
  \multirow{1}{*}{IDHN-GoogLeNet} & 0.5742 & 0.5801 & 0.5925 & 0.5984 & \textbf{0.5603} & \textbf{0.5738} & 0.5807 & \textbf{0.5876} & 0.7890 & 0.7972 &.8030 & 0.8050 \\
  %\hline
  \multirow{1}{*}{IDHN-VGG19} & \textbf{0.5892} & \textbf{0.6161} & \textbf{0.6211} & \textbf{0.6238} & 0.5600 & 0.5692 & \textbf{0.5834} & 0.5845 & \textbf{0.7949} & \textbf{0.8082} & \textbf{0.8139} & \textbf{0.8201} \\
  \hline
\end{tabular}}
\label{table:four}
\end{table*}

Similarly, Figure~\ref{fig:flickr25k} and \ref{fig:voc2012} show the ACG, NDCG and precision curves on the Flickr and VOC2012, respectively. It can be seen that, the proposed method achieves the highest performance on the two datasets. On Flickr, the proposed IDHN obtains significantly higher performance than the other methods. On VOC2012, among the comparison methods, HashNet achieves the most comparative results to IDHN, especially on 12 bits and 24 bits. However, when the hash code increases to 36 bits or 48 bits, the proposed IDHN holds an obvious higher performance than HashNet.

According to the definition of MAP, pairwise images that share at least one common object label will be considered as relevant images, and no more comparisons of fine-grained semantic relation between these images are included, which may not stay in step with user demand in multi-label image retrieval. We hope that high-quality retrieval results should have as more shared class labels as possible in the nearest retrieval image. Therefore, we also use WAP to measure the average number of shared class labels among these retrieved similar images. Figure~\ref{fig:wap} show the results of WAP for different numbers of bits. It can be seen that, although the range of WAP on the three datasets are very different, the WAPs of IDHN are stably higher than that of the comparison methods.

\begin{figure*}[t!]
  \centering
  \centerline{\includegraphics[width=1.0\linewidth]{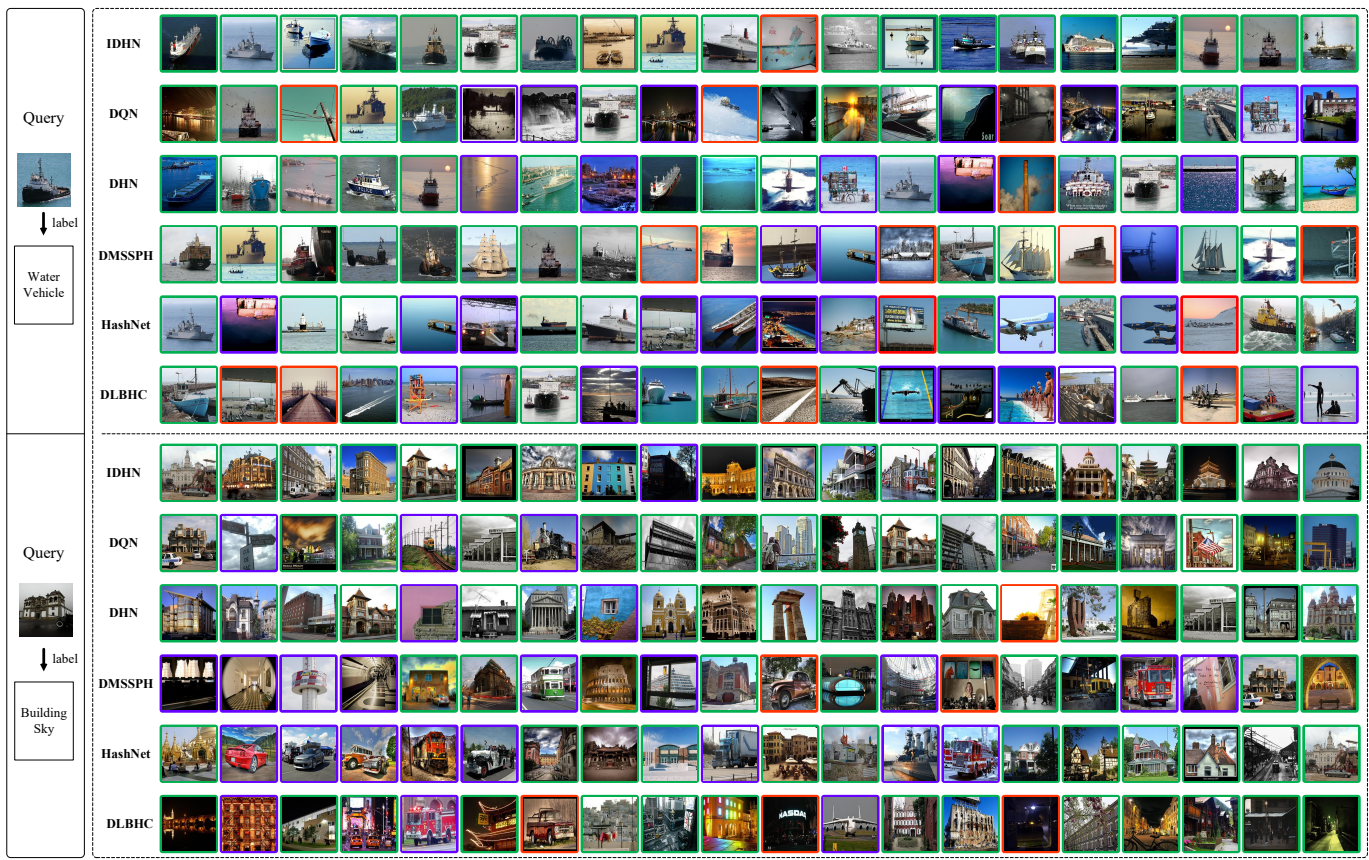}}
\caption{Top 20 retrieved images of the proposed IDHN and five competing deep hashing methods using the Hamming ranking on 48-bit hash codes. The green box indicates that the retrieved image contains all the object classes in the query image, the blue box indicates the retrieved image partially contains the target classes, and the red box indicates the retrieved image does not contain any object classes in the query image.}
\label{fig:retrieval}
\end{figure*}

We also compare the proposed method with another four deep hashing methods, DPSH, DSDH, DSRH and DTSH. According to the results in Table~\ref{table:compare}, the proposed method IDHN outperforms the two pairwise-similarity-based methods DPSH and DSDH with a significant margin. It indicates that, the soft similarity is helpful for learning high-quality hash codes. IDHN also achieves more robust results than the two triplet-based methods DSRH and DTSH on the four datasets at different code lengths.

\subsubsection{Comparison with different settings}
To justify the necessary of using the soft similarity definition and joint loss function, we conduct some comparison experiments.
Table~\ref{table:four} shows the results of MAP, WAP, ACG and NDCG metrics of the IDHN and its modifications, respectively.
IDHN-fine-ce and IDHN-fine-mse are both trained under supervision of our proposed pairwise quantified similarity, where the difference between them is that IDHN-fine-ce only uses the cross entropy loss and IDHN-fine-mse only uses mean square error loss. As a contrast, IDHN-coarse-ce and IDHN-coarse-mse are trained under supervision of coarse pairwise similarity that is adapted by most deep hashing methods at present. IDHN-coarse-ce only uses the cross entropy loss and IDHN-coarse-mse only uses mean square error loss.

Comparing IDHN-fine-ce with IDHN-coarse-ce, or IDHN-fine-mse with IDHN-coarse-mse, we can observe that using the pairwise quantified similarity as the supervised information can achieve results with higher semantic similarity than that obtained by the coarse similarity definition, whether adapting the cross entropy loss or mean square error loss. To some extent, the value of WAP and ACG can reflect the degree of shared labels between multi-label images. That is, the larger the value the larger the average number of shared labels over the whole dataset. It also means that, there are more complex semantic relation between images. Although IDHN-fine-ce has achieved outstanding performance on VOC2102 dataset, it does not show advantage on other two datasets -- NUS-WIDE and Flickr, while there are complicated and abundant semantic relation on these two image datasets than VOC2012. It is interesting that with more fine-grained similarity information, deep hashing model using mean-square-error loss presents comparable performance compared to deep hashing model using cross-entropy loss, especially on dataset of complex semantic relations. The proposed IDHN method combines the advantages of cross entropy loss and mean square error loss, and shows sufficient powerful and robust performance on the three multi-label image datasets.

For fair comparison, all the above experiments are conducted based on AlexNet. We also extend our method to other deep network bases -- GoogLeNet and VGG19, both of which achieve more accurate results than AlexNet on the ImageNet competition, to explore the extensibility of our hashing strategy. We denote these two modifications as `IDHN-GoogLeNet' and `IDHN-VGG19', respectively. From Table~\ref{table:four} we can see that, with more powerful network bases, IDHN achieves improved performance on all these metrics, which indicates a good transfer capability of the proposed deep hashing strategy.

\subsubsection{Top retrieval results}
Figure~\ref{fig:retrieval} shows some retrieval samples of six deep learning methods according to the ascending Hamming ranking. We mark  the retrieval image with green box that includes all object classes in query image, blue box that includes part of the object classes, and red box which does not include any object classes in the query image. The first query image contains two semantic labels: water and vehicle. We can see that, among these six deep hashing methods, IDHN shows the best suitability between the retrieval images and query images, since the IDHN has the least incorrect retrieval (marked by red box) in top-20 retrieval results. The second query images contains two semantic labels: building and sky. On the top-20 retrieval images of each method, the results of IDHN are more similar to query images from the perspective of human vision. The top-20 results of DHN, DMSSPH and DLBHC have incorrect retrieval results, and the top-20 results of DQN and HashNet include some low similarity results, as denoted by the blue-box images. The results show the advantage of the proposed method for multi-label image retrieval.

\section{Conclusion}
\label{sec:conc}
In this paper, a novel deep hashing method was proposed for multi-label image retrieval, in which a quantified similarity definition was introduced to measure the fine-grained pairwise similarity. Compared with the traditional pairwise similarity, this fine-grained pairwise similarity can more effectively encode the information of fine-grained multi-label images. Based on the proposed pairwise similarity, a robust pairwise-similarity loss combining the cross-entropy loss and the mean-square-error loss  was constructed for effective similarity-preserving learning. In addition, a quantization loss was introduced to control the quality of hashing. Extensive experiments on four multi-label datasets demonstrated that, the proposed IDHN outperformed the competing methods and achieved an effective feature learning and hash-code learning in the multi-label image retrieval.

\bibliographystyle{IEEEtran}
\bibliography{deephash}

\begin{thebibliography}{10}
\providecommand{\url}[1]{#1}
\csname url@samestyle\endcsname
\providecommand{\newblock}{\relax}
\providecommand{\bibinfo}[2]{#2}
\providecommand{\BIBentrySTDinterwordspacing}{\spaceskip=0pt\relax}
\providecommand{\BIBentryALTinterwordstretchfactor}{4}
\providecommand{\BIBentryALTinterwordspacing}{\spaceskip=\fontdimen2\font plus
\BIBentryALTinterwordstretchfactor\fontdimen3\font minus
  \fontdimen4\font\relax}
\providecommand{\BIBforeignlanguage}[2]{{%
\expandafter\ifx\csname l@#1\endcsname\relax
\typeout{** WARNING: IEEEtran.bst: No hyphenation pattern has been}%
\typeout{** loaded for the language `#1'. Using the pattern for}%
\typeout{** the default language instead.}%
\else
\language=\csname l@#1\endcsname
\fi
#2}}
\providecommand{\BIBdecl}{\relax}
\BIBdecl

\bibitem{tao2006direct}
D.~Tao, X.~Tang, X.~Li, and Y.~Rui, ``Direct kernel biased discriminant
  analysis: a new content-based image retrieval relevance feedback algorithm,''
  \emph{IEEE Transactions on Multimedia}, vol.~8, no.~4, pp. 716--727, 2006.

\bibitem{wang2014hashing}
J.~Wang, H.~T. Shen, J.~Song, and J.~Ji, ``Hashing for similarity search: A
  survey,'' \emph{arXiv preprint arXiv:1408.2927}, 2014.

\bibitem{weiss2009spectral}
Y.~Weiss, A.~Torralba, and R.~Fergus, ``Spectral hashing,'' in \emph{Advances
  in neural information processing systems}, 2009, pp. 1753--1760.

\bibitem{kulis2009learning}
B.~Kulis and T.~Darrell, ``Learning to hash with binary reconstructive
  embeddings,'' in \emph{Advances in neural information processing systems},
  2009, pp. 1042--1050.

\bibitem{wang2010semi}
J.~Wang, S.~Kumar, and S.-F. Chang, ``Semi-supervised hashing for scalable
  image retrieval,'' in \emph{IEEE Conference on Computer Vision and Pattern
  Recognition}, 2010, pp. 3424--3431.

\bibitem{norouzi2011minimal}
M.~Norouzi and D.~M. Blei, ``Minimal loss hashing for compact binary codes,''
  in \emph{International Conference on Machine Learning}, 2011, pp. 353--360.

\bibitem{liu2012supervised}
W.~Liu, J.~Wang, R.~Ji, Y.-G. Jiang, and S.-F. Chang, ``Supervised hashing with
  kernels,'' in \emph{IEEE Conference on Computer Vision and Pattern
  Recognition}, 2012, pp. 2074--2081.

\bibitem{gong2013iterative}
Y.~Gong, S.~Lazebnik, A.~Gordo, and F.~Perronnin, ``Iterative quantization: A
  procrustean approach to learning binary codes for large-scale image
  retrieval,'' \emph{IEEE Transactions on Pattern Analysis and Machine
  Intelligence}, vol.~35, no.~12, pp. 2916--2929, 2013.

\bibitem{Li2015Ordinal}
C.~Li, Q.~Liu, J.~Liu, and H.~Lu, ``Ordinal distance metric learning for image
  ranking,'' \emph{IEEE Transactions on Neural Networks and Learning Systems},
  vol.~26, no.~7, p. 1551, 2015.

\bibitem{Liu2016Sequential}
L.~Liu and L.~Shao, ``Sequential compact code learning for unsupervised image
  hashing,'' \emph{IEEE Transactions on Neural Networks and Learning Systems},
  vol.~27, no.~12, pp. 2526--2536, 2016.

\bibitem{Jie2016Supervised}
G.~Jie, T.~Liu, Z.~Sun, D.~Tao, and T.~Tan, ``Supervised discrete hashing with
  relaxation,'' \emph{IEEE Transactions on Neural Networks and Learning
  Systems}, vol.~PP, no.~99, pp. 1--10, 2017.

\bibitem{Liu2017Reversed}
Q.~Liu, G.~Liu, L.~Li, X.~T. Yuan, M.~Wang, and W.~Liu, ``Reversed spectral
  hashing,'' \emph{IEEE Transactions on Neural Networks and Learning Systems},
  vol.~PP, no.~99, pp. 1--9, 2017.

\bibitem{shen2018unsupervised}
F.~Shen, Y.~Xu, L.~Liu, Y.~Yang, Z.~Huang, and H.~T. Shen, ``Unsupervised deep
  hashing with similarity-adaptive and discrete optimization,'' \emph{IEEE
  transactions on Pattern Analysis and Machine Intelligence}, vol.~40, no.~12,
  pp. 3034--3044, 2018.

\bibitem{zhou2018graph}
X.~Zhou, F.~Shen, L.~Liu, W.~Liu, L.~Nie, Y.~Yang, and H.~T. Shen, ``Graph
  convolutional network hashing,'' \emph{IEEE Transactions on Cybernetics}, pp.
  1--13, 2018.

\bibitem{krizhevsky2012imagenet}
A.~Krizhevsky, I.~Sutskever, and G.~E. Hinton, ``Imagenet classification with
  deep convolutional neural networks,'' in \emph{Advances in neural information
  processing systems}, 2012, pp. 1097--1105.

\bibitem{simonyan2014very}
K.~Simonyan and A.~Zisserman, ``Very deep convolutional networks for
  large-scale image recognition,'' \emph{arXiv preprint arXiv:1409.1556}, 2014.

\bibitem{szegedy2015going}
C.~Szegedy, W.~Liu, Y.~Jia, P.~Sermanet, S.~Reed, D.~Anguelov, D.~Erhan,
  V.~Vanhoucke, and A.~Rabinovich, ``Going deeper with convolutions,'' in
  \emph{IEEE Conference on Computer Vision and Pattern Recognition}, 2015, pp.
  1--9.

\bibitem{grsl2015}
Q.~{Zou}, L.~{Ni}, T.~{Zhang}, and Q.~{Wang}, ``Deep learning based feature
  selection for remote sensing scene classification,'' \emph{IEEE Geoscience
  and Remote Sensing Letters}, vol.~12, no.~11, pp. 2321--2325, 2015.

\bibitem{szegedy2013deep}
C.~Szegedy, A.~Toshev, and D.~Erhan, ``Deep neural networks for object
  detection,'' in \emph{Advances in neural information processing systems},
  2013, pp. 2553--2561.

\bibitem{sun2014deep}
Y.~Sun, Y.~Chen, X.~Wang, and X.~Tang, ``Deep learning face representation by
  joint identification-verification,'' in \emph{Advances in neural information
  processing systems}, 2014, pp. 1988--1996.

\bibitem{long2015fully}
J.~Long, E.~Shelhamer, and T.~Darrell, ``Fully convolutional networks for
  semantic segmentation,'' in \emph{IEEE Conference on Computer Vision and
  Pattern Recognition}, 2015, pp. 3431--3440.

\bibitem{deng2014large}
J.~Deng, N.~Ding, Y.~Jia, A.~Frome, K.~Murphy, S.~Bengio, Y.~Li, H.~Neven, and
  H.~Adam, ``Large-scale object classification using label relation graphs,''
  in \emph{European Conference on Computer Vision}, 2014, pp. 48--64.

\bibitem{Zou2018deepcrack}
Q.~Zou, Z.~Zhang, Q.~Li, X.~Qi, Q.~Wang, and S.~Wang, ``Deepcrack: Learning
  hierarchical convolutional features for crack detection,'' \emph{IEEE
  Transactions on Image Processing}, vol.~28, no.~3, pp. 1498--1512, 2019.

\bibitem{xia2014supervised}
R.~Xia, Y.~Pan, H.~Lai, C.~Liu, and S.~Yan, ``Supervised hashing for image
  retrieval via image representation learning.'' in \emph{AAAI Conference on
  Artificial Intelligence}, vol.~1, 2014, pp. 2156--2162.

\bibitem{zhao2015deep}
F.~Zhao, Y.~Huang, L.~Wang, and T.~Tan, ``Deep semantic ranking based hashing
  for multi-label image retrieval,'' in \emph{IEEE Conference on Computer
  Vision and Pattern Recognition}, 2015, pp. 1556--1564.

\bibitem{lai2015simultaneous}
H.~Lai, Y.~Pan, Y.~Liu, and S.~Yan, ``Simultaneous feature learning and hash
  coding with deep neural networks,'' in \emph{IEEE Conference on Computer
  Vision and Pattern Recognition}, 2015, pp. 3270--3278.

\bibitem{zhang2015bit}
R.~Zhang, L.~Lin, R.~Zhang, W.~Zuo, and L.~Zhang, ``Bit-scalable deep hashing
  with regularized similarity learning for image retrieval and person
  re-identification,'' \emph{IEEE Transactions on Image Processing}, vol.~24,
  no.~12, pp. 4766--4779, 2015.

\bibitem{zhu2016deep}
H.~Zhu, M.~Long, J.~Wang, and Y.~Cao, ``Deep hashing network for efficient
  similarity retrieval.'' in \emph{AAAI Conference on Artificial Intelligence},
  2016, pp. 2415--2421.

\bibitem{cao2016deep}
Y.~Cao, M.~Long, J.~Wang, H.~Zhu, and Q.~Wen, ``Deep quantization network for
  efficient image retrieval.'' in \emph{AAAI Conference on Artificial
  Intelligence}, 2016, pp. 3457--3463.

\bibitem{liu2016deep}
H.~Liu, R.~Wang, S.~Shan, and X.~Chen, ``Deep supervised hashing for fast image
  retrieval,'' in \emph{IEEE Conference on Computer Vision and Pattern
  Recognition}, 2016, pp. 2064--2072.

\bibitem{cao2017hashnet}
Z.~Cao, M.~Long, J.~Wang, and P.~S. Yu, ``Hashnet: Deep learning to hash by
  continuation,'' \emph{arXiv preprint arXiv:1702.00758}, 2017.

\bibitem{li2017deep}
Q.~Li, Z.~Sun, R.~He, and T.~Tan, ``Deep supervised discrete hashing,''
  \emph{arXiv preprint arXiv:1705.10999}, 2017.

\bibitem{wang2016deep}
X.~Wang, Y.~Shi, and K.~M. Kitani, ``Deep supervised hashing with triplet
  labels,'' in \emph{Asian conference on computer vision}, 2016, pp. 70--84.

\bibitem{li2016feature}
W.~Li, S.~Wang, and W.~Kang, ``Feature learning based deep supervised hashing
  with pairwise labels,'' \emph{International Joint Conference on Artificial
  Intelligence}, pp. 1711--1717, 2016.

\bibitem{jiang2017deep}
Q.-Y. Jiang and W.-J. Li, ``Deep cross-modal hashing,'' in \emph{IEEE
  conference on computer vision and pattern recognition}, 2017, pp. 3232--3240.

\bibitem{he2018hashing}
K.~He, F.~Cakir, S.~Adel~Bargal, and S.~Sclaroff, ``Hashing as tie-aware
  learning to rank,'' in \emph{IEEE Conference on Computer Vision and Pattern
  Recognition}, 2018, pp. 4023--4032.

\bibitem{datar2004locality}
M.~Datar, N.~Immorlica, P.~Indyk, and V.~S. Mirrokni, ``Locality-sensitive
  hashing scheme based on p-stable distributions,'' in \emph{Annual Symposium
  on Computational Geometry}, 2004, pp. 253--262.

\bibitem{li2013spectral}
P.~Li, M.~Wang, J.~Cheng, C.~Xu, and H.~Lu, ``Spectral hashing with
  semantically consistent graph for image indexing,'' \emph{IEEE Transactions
  on Multimedia}, vol.~15, no.~1, pp. 141--152, 2013.

\bibitem{ning2017scalable}
Q.~Ning, J.~Zhu, Z.~Zhong, S.~C.~H. Hoi, and C.~Chen, ``Scalable image
  retrieval by sparse product quantization,'' \emph{IEEE Transactions on
  Multimedia}, vol.~19, no.~3, pp. 586--597, 2017.

\bibitem{ercoli2017compact}
S.~Ercoli, M.~Bertini, and A.~D. Bimbo, ``Compact hash codes for efficient
  visual descriptors retrieval in large scale databases,'' \emph{IEEE
  Transactions on Multimedia}, vol.~19, no.~11, pp. 2521--2532, 2017.

\bibitem{lu2016latent}
X.~Lu, X.~Zheng, and X.~Li, ``Latent semantic minimal hashing for image
  retrieval,'' \emph{IEEE Transactions on Image Processing}, 2016.

\bibitem{Huang2017Online}
L.~K. Huang, Q.~Yang, and W.~S. Zheng, ``Online hashing,'' \emph{IEEE
  Transactions on Neural Networks and Learning Systems}, vol.~PP, no.~99, pp.
  1--14, 2017.

\bibitem{Xia2013Online}
H.~Xia, S.~C.~H. Hoi, R.~Jin, and P.~Zhao, ``Online multiple kernel similarity
  learning for visual search,'' \emph{IEEE Transactions on Pattern Analysis and
  Machine Intelligence}, vol.~36, no.~3, pp. 536--549, 2013.

\bibitem{Liang2017Semisupervised}
J.~Liang, Q.~Hu, W.~Wang, and Y.~Han, ``Semisupervised online multikernel
  similarity learning for image retrieval,'' \emph{IEEE Transactions on
  Multimedia}, vol.~19, no.~5, pp. 1077--1089, 2017.

\bibitem{zou2017robust}
Q.~Zou, L.~Ni, Q.~Wang, Q.~Li, and S.~Wang, ``Robust gait recognition by
  integrating inertial and rgbd sensors,'' \emph{IEEE Transactions on
  Cybernetics}, vol.~48, no.~4, pp. 1136--1150, 2018.

\bibitem{Lin2015Deep}
K.~Lin, H.~F. Yang, J.~H. Hsiao, and C.~S. Chen, ``Deep learning of binary hash
  codes for fast image retrieval,'' in \emph{Computer Vision and Pattern
  Recognition Workshops}, 2015, pp. 27--35.

\bibitem{lai2016instance}
H.~Lai, P.~Yan, X.~Shu, Y.~Wei, and S.~Yan, ``Instance-aware hashing for
  multi-label image retrieval,'' \emph{IEEE Transactions on Image Processing},
  vol.~25, no.~6, pp. 2469--2479, 2016.

\bibitem{wu2017deep}
D.~Wu, Z.~Lin, B.~Li, M.~Ye, and W.~Wang, ``Deep supervised hashing for
  multi-label and large-scale image retrieval,'' in \emph{International
  Conference on Multimedia Retrieval}, 2017, pp. 150--158.

\bibitem{baeza1999modern}
R.~Baeza-Yates, B.~Ribeiro-Neto \emph{et~al.}, \emph{Modern information
  retrieval}.\hskip 1em plus 0.5em minus 0.4em\relax ACM press New York, 1999,
  vol. 463.

\bibitem{jarvelin2002cumulated}
K.~Jarvelin and J.~Kekalainen, ``Cumulated gain-based evaluation of {IR}
  techniques,'' \emph{ACM Transactions on Information Systems}, vol.~20, no.~4,
  pp. 422--446, 2002.

\bibitem{sablayrolles2017icassp}
A.~Sablayrolles, M.~Douze, N.~Usunier, and H.~J{\'e}gou, ``How should we
  evaluate supervised hashing?'' in \emph{IEEE International Conference on
  Acoustics, Speech and Signal Processing (ICASSP)}, 2017, pp. 1732--1736.

\bibitem{chua2009nus}
T.-S. Chua, J.~Tang, R.~Hong, H.~Li, Z.~Luo, and Y.~Zheng, ``Nus-wide: a
  real-world web image database from national university of singapore,'' in
  \emph{ACM International Conference on Image and Video Retrieval}, 2009.

\bibitem{liu2011hashing}
W.~Liu, J.~Wang, S.~Kumar, and S.-F. Chang, ``Hashing with graphs,'' in
  \emph{International Conference on Machine Learning}, 2011, pp. 1--8.

\bibitem{huiskes2008mir}
M.~J. Huiskes and M.~S. Lew, ``The mir flickr retrieval evaluation,'' in
  \emph{ACM International Conference on Multimedia Information Retrieval},
  2008, pp. 39--43.

\bibitem{everingham2010pascal}
M.~Everingham, L.~Van~Gool, C.~K. Williams, J.~Winn, and A.~Zisserman, ``The
  pascal visual object classes (voc) challenge,'' \emph{International journal
  of computer vision}, vol.~88, no.~2, pp. 303--338, 2010.

\bibitem{escalante2010the}
H.~J. Escalante, C.~A. Hernandez, J.~A. Gonzalez, A.~Lopezlopez, M.~Montes,
  E.~F. Morales, L.~E. Sucar, L.~Villasenor, and M.~Grubinger, ``The segmented
  and annotated iapr tc-12 benchmark,'' \emph{Computer Vision and Image
  Understanding}, vol. 114, no.~4, pp. 419--428, 2010.

\bibitem{schroff2015facenet}
F.~Schroff, D.~Kalenichenko, and J.~Philbin, ``Facenet: A unified embedding for
  face recognition and clustering,'' in \emph{IEEE Conference on Computer
  Vision and Pattern Recognition}, 2015, pp. 815--823.

\bibitem{abadi2016tensorflow}
M.~Abadi, A.~Agarwal, P.~Barham, E.~Brevdo, Z.~Chen, C.~Citro, G.~S. Corrado,
  A.~Davis, J.~Dean, M.~Devin \emph{et~al.}, ``Tensorflow: Large-scale machine
  learning on heterogeneous distributed systems,'' \emph{arXiv preprint
  arXiv:1603.04467}, 2016.

\bibitem{sis}
Y.~Yuan, ``Habir: Hashing baseline for image retrieval,''
  \url{https://github.com/willard-yuan/hashing-baseline-for-image-retrieval}.

\bibitem{jarvelin2000ir}
K.~Jarvelin and J.~Kekalainen, ``{IR} evaluation methods for retrieving highly
  relevant documents,'' in \emph{ACM SIGIR Conference on Research and
  Development in Information Retrieval}, 2000, pp. 41--48.

\end{thebibliography}

\end{document}